\documentclass{article}

\usepackage{arxiv}

\usepackage[utf8]{inputenc} 
\usepackage[T1]{fontenc}    
\usepackage{hyperref}       
\usepackage{amsmath}
\usepackage{url}            
\usepackage{booktabs}       
\usepackage{amsfonts}       
\usepackage{nicefrac}       
\usepackage{microtype}      
\usepackage{lipsum}		
\usepackage{graphicx}
\usepackage{natbib}
\usepackage{algorithm}
\usepackage{amsmath}
\usepackage{graphicx}
\usepackage[varg]{txfonts}
\usepackage{bm}
\usepackage{float}
\usepackage{threeparttable}
\usepackage{algorithm}
\usepackage{algpseudocode}
\usepackage{subfigure}
\usepackage{epstopdf}
\usepackage{booktabs}
\usepackage{multirow}
\usepackage{color}
\usepackage{amsfonts,amssymb,array}
\usepackage{hyperref}
\usepackage{doi}

\title{PTR-PPO: Proximal Policy Optimization with Prioritized Trajectory Replay\thanks{Yanghe Feng  and Zhong Liu are the Co-corresponding author.}\thanks{ Xingxing Liang and Yang Ma are equal contribution.}}

\author{ {\hspace{1mm}Xingxing Liang} \\
	College of Systems Engineering\\
	National University of Defense Technology\\
	Changsha, CN 410072 \\
	\texttt{doublestar\_l@163.com} \\
	\And
	{\hspace{1mm}Yang Ma} \\
	College of Systems Engineering\\
	National University of Defense Technology\\
	Changsha, CN 410072 \\
	\texttt{yang\_ma\_cn@163.com} \\
	\And
	{\hspace{1mm}Yanghe Feng} \\
	College of Systems Engineering\\
	National University of Defense Technology\\
	Changsha, CN 410072 \\
	\texttt{fengyanghe@nudt.edu.cn} \\
	\And
	{\hspace{1mm}Zhong Liu} \\
	College of Systems Engineering\\
	National University of Defense Technology\\
	Changsha, CN 410072 \\
	\texttt{phillipliu@263.net} \\
}

\renewcommand{\shorttitle}{PTR-PPO}

\hypersetup{
pdftitle={PTR-PPO: Proximal Policy Optimization with Prioritized Trajectory Replay},
pdfsubject={Computer Science, Machine Learning},
pdfauthor={Xingxing Liang},
pdfkeywords={deep reinforcement learning, proximal policy optimization, prioritized experience replay, temporal-difference error, Atari},
}

\begin{document}
\maketitle

\begin{abstract}
	On-policy deep reinforcement learning algorithms have low data utilization and require significant experience for policy improvement. This paper proposes a proximal policy optimization algorithm with prioritized trajectory replay (PTR-PPO) that combines on-policy and off-policy methods to improve sampling efficiency by prioritizing the replay of trajectories generated by old policies. We first design three trajectory priorities based on the characteristics of trajectories: the first two being max and mean trajectory priorities based on one-step empirical generalized advantage estimation (GAE) values and the last being reward trajectory priorities based on normalized undiscounted cumulative reward. Then, we incorporate the prioritized trajectory replay into the PPO algorithm, propose a truncated importance weight method to overcome the high variance caused by large importance weights under multistep experience, and design a policy improvement loss function for PPO under off-policy conditions. We evaluate the performance of PTR-PPO in a set of Atari discrete control tasks, achieving state-of-the-art performance. In addition, by analyzing the heatmap of priority changes at various locations in the priority memory during training, we find that memory size and rollout length can have a significant impact on the distribution of trajectory priorities and, hence, on the performance of the algorithm.
\end{abstract}


\section{Introduction}
Deep reinforcement learning (DRL) enables agents to improve themselves through trial-and-error learning and has achieved good results in fields such as Go \citep{silver2016mastering,silver2017mastering,schrittwieser2020mastering}, games \citep{mnih2015human,vinyals2019grandmaster}, and automatic control \citep{duan2020hierarchical,liang2020deep}. A good simulation environment can generate a large amount of experience for agent learning through rapid high-fidelity simulation, which is one of the key elements for the success of DRL. However, limited by time and space, hardware and software requirements, the real system cannot accelerate and generate a large number of interactions with the agent. It is difficult to generate a large quantity of data for online agent learning, and there is an urgent need to improve the sample efficiency of DRL algorithms. Existing on-policy DRL methods\citep{mnih2016asynchronous,schulman2017proximal}, which cannot use past generation data, give up some rare experience that needs to be replayed for learning that will be useful during training. In off-policy DRL methods\citep{mnih2015human}, to enable the agent to remember and reuse past experience and improve sample efficiency, experience replay mechanisms \citep{mnih2015human,wang2016sample,lin1992self} are used to store experience generated from interactions with the environment, and this experience is sampled evenly during training. However, in the vast majority of experience updates, the state values learned by the agent are simply transferred from one initial state value to another, especially in reward-sparse environments, where only those experiences that jump into or out of the state prior to the endpoint goal result in a change in the value estimate\citep{sutton2018reinforcement}. In the case of uniform sampling, many useless update calculations are performed before a useful experience sample is collected, which leads to extremely low sample efficiency in an environment with a large state space.

Intuitively, experience can be prioritized according to importance, and this approach increases the replay probability of important experiences, thereby making learning more efficient. Prioritized experience replay (PER) \citep{schaul2015prioritized,horgan2018distributed} makes the agent preferentially select samples for experience replay and frequently replays those experience samples with high expectations that use temporal-difference error\citep{sutton2018reinforcement} to measure the level of priority. As training updates increase, the estimated distribution of values will have bias with respect to the true distribution, and PER reduces the influence of the problem on the training results by introducing importance-sample weights (IS weights) to correct the distribution.

PER provides a better empirical convergence guarantee for the convergence of reinforcement learning algorithms, deriving various RL algorithms in combination with PER: PER-DQN \citep{horgan2018distributed}, PER-DDPG\citep{hou2017novel}, PER-HER\citep{zhao2018energy}, PER-NAF\citep{gu2016continuous}, PER-TD3\citep{fujimoto2018addressing}, etc. Moreover, the original PER was applied only from the sampling perspective, and some people\citep{daley2019reconciling,bu2020double} have used PER in the experience storage phase as well, focusing more on the preservation of important experiences.

\begin{figure*}[!ht]
\centering
\includegraphics[width=14.46cm,height=6.18cm]{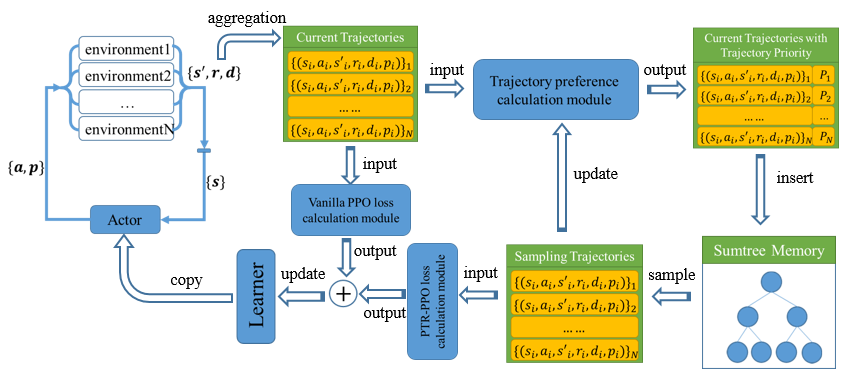}
\caption{PTR-PPO Algorithm Architecture. We use the learner-actor architecture to train the agent. The top-left part shows the actor interacting with multiple environments in parallel, generating multiple experiences, aggregating these experiences to obtain the \emph{current trajectories}, using the \emph{trajectory preference calculation module} to obtain the priority of each trajectory, inserting them in the \emph{sumtree memory} in the priority memory according to the trajectory priority, and sampling the experiences from the \emph{sumtree memory} to obtain \emph{sampling trajectories}. Afterward, these trajectories are input into the \emph{PTR-PPO loss calculation module} to obtain PTR-PPO loss; \emph{current trajectories} are input into the \emph{vanilla PPO loss calculation module} to obtain the ppo loss. Both losses are backpropagated and the learner parameters are updated, after which the learner passes the parameters to the actor to generate new trajectories.}
\label{fig:arch}
\end{figure*}

However, existing DRL algorithms that combine with PER are off-policy methods. On-policy DRL-based policy gradient (PG) methods \citep{mnih2016asynchronous,schulman2017proximal,gu2016continuous,schulman2015trust}, which are popular in deep reinforcement learning research, have extremely low sampling efficiency, which severely limits the algorithm performance. In addition, existing PER methods measure the importance of one-step experience. However, updating state values using one-step experience leads to large bias, especially in reward-sparse environments, and using multistep experience for state value evaluation is one of the advantages of on-policy DRL method pairs.

How to use the priority replay mechanism of multistep experience is an important factor to improve the sample efficiency of on-policy PG methods. Among PG methods, proximal policy optimization (PPO) \citep{schulman2017proximal}, which uses multistep experience for value evaluation and makes the policy steadily improve by limiting the gap between the old and new policy, is an advanced deep reinforcement learning algorithm.

In this paper, we propose a new reinforcement learning algorithm, called proximal policy optimization with prioritized trajectory replay (PTR-PPO), to improve the learning speed of the RL algorithm by improving sample efficiency. The main contributions of this paper are as follows:

1) The obtained trajectory performance is analyzed, and two trajectory priority metrics are proposed: the generalized advantage estimator (GAE) \citep{schulman2017proximal,schulman2015high,kabir2021effects}-based trajectory priority metric and the trajectory priority metric based on the trajectory undiscounted cumulative reward. Compared with the original temporal-difference error, GAE effectively reduces the estimation bias of experiences in trajectory and is a technique used by the current mainstream reinforcement learning algorithms \citep{schulman2017proximal}. According to GAE, the trajectory priority based on the max factor of one-step experience advantage bias and the trajectory priority based on the mean factor of one-step experience advantage bias are designed to measure the priority of the trajectory. According to the trajectory undiscounted cumulative reward, the trajectory priority of the normalized undiscounted cumulative reward is designed to improve the replay probability of experience with significant reward signal and overcome the sparse reward.

2) Based on the trajectory priority, the proximal policy optimization with a prioritized trajectory replay algorithm and framework are proposed to reuse the past important trajectories. PTR-PPO combines the general PPO training process with the prioritized trajectory replay method, and the sampled trajectories and current trajectories are used to jointly update the policy, which improves the sample efficiency and performance of the PPO algorithm.

3) Experiments on Atari benchmarks demonstrate that the proposed PTR-PPO algorithm outperforms or matches PPO \citep{schulman2017proximal} and ACER \citep{wang2016sample} across all benchmark tasks in terms of the final performance.

The main structure of the paper is as follows. In Section II, we describe some preliminaries of RL. Section III designs three trajectory priority factors. In Section IV, we present the PTR-PPO architecture and algorithm. In Section V, presents experimental results that show the efficacy of PTR-PPO. Section VI concludes this paper.

\section{Background}
\subsection{Deep Reinforcement Learning}

Reinforcement learning aims to find the policy with the maximum cumulative reward in a sequential decision task\citep{liang2020novel}. In a reinforcement learning task, the agent interacts with the environment over time. At each time step $t$, the agent receives a state $s_{t}$ in state space $ \mathcal{S}$ and, according to the policy $\pi(\cdot|{s_{t}})$, selects an action $a_{t}$ from the action space $ \mathcal{A}$ to interact with the environment, that is the agent's behavior, after which the environment returns a scalar reward value $r_{t}$ and the next state $s_{t+1}$ according to the dynamics of the environment (reward function $\mathcal{R}(s,a,s')$ and state transfer function $\mathcal{P}(s_{t+1}|{s_{t},a_{t}})$). A trial is conducted until the agent reaches the termination state, after which it proceeds to the next trial. The return $R_{t}=\sum_{k=1}^{\infty }{{\gamma }^{k-1}}{{r}_{t+k}}$is a cumulative reward with a discount $\gamma \in \left( 0,1 \right]$, and the agent's goal is to maximize the return expectation for each state.

With the development of deep learning, neural networks are widely used  to improve the performance of reinforcement learning. Deep Q learning (DQN) \citep{mnih2015human} uses a convolutional and feed-forward network with parameter $\theta$ to learn the action value function $Q_{\theta}(s,a)$. The value network accepts a picture representation of the state $s_{t}$ and outputs a vector containing the value of every action $a_{t}$. DQN interacts with the environment in a greedy manner $\epsilon-greedy$ by selecting the action with the highest value from the action value estimate and cannot give a preference for actions with similar action values. The asynchronous advantage actor-critic algorithm (A3C)\citep{mnih2016asynchronous} consists of two networks, actor $\pi(a|s,\theta_{\pi})$and critic $V(s,\theta_{v})$, which use a similar state encoding network architecture as DQN, with the former outputting the probability of each action and the latter outputting the state value. A3C is an on-policy RL method, that improves the training sample size and reduces the learning time by an asynchronous approach, and uses a multistep empirical estimation value function, which has a high learning efficiency.

\subsection{Experience Replay}

Experience replay\citep{sutton2018reinforcement} uniformly samples experience from replay memory, addressing the limitation that an agent learning online must learn strictly in the order of experienced samples. PER is a widely used technique in experience replay that enables an agent to preferentially select samples for experience replay learning by increasing the sampling probability of important samples.

PER replays those experience samples with high expectations more frequently and measures the priority by TD error and its variants. The TD-error variant in the PER-DQN algorithm is calculated in Equation\ref {eq:1}.

\begin{equation}\label{eq:1}
  |\delta|=|r+\gamma Q(s',a';\theta')-Q(s,a;\theta)|
\end{equation}
where $\theta$ and $\theta'$ represent the value and target value network parameters, respectively.

By adjusting the parameters that combine PER with uniform sampling, the sharpness of the priority can be weakened; then, the sampling probability of a transition becomes:
\begin{equation}\label{eq:2}
  P(i)=\frac{p_{i}^{\alpha}}{\sum_{k}p_{k}^{\alpha}}
\end{equation}

where $p_{i}$ is the priority of sample $i$, which is greater than 0; the exponential $\alpha$ is a parameter that determines the priority level, and the case $\alpha =0$ corresponds to a random sampling method with equal probability. For the definition of the priority value $p_{i}$, there are two variants: the first is proportional prioritization, calculated as $p_i=|\delta_i|+\epsilon$, where $\epsilon$ is a small positive number, to enable some special marginal samples with a TD-error value of 0 to be sampled; the second is prioritization based on sorting, where $p_i=\frac{1}{rank(i)}$, where $rank(i)$ is determined according to the rank $|\delta_i|$of the sample $i$ in sorting. For both variants, $P(i)$ is monotonic with respect to $|\delta_{i}|$. The second variant, may be more robust due to its insensitivity to outliers. Both variants can accelerate the convergence of the algorithm, but the first variant is most commonly used in practice.

\subsection{Proximal Policy Optimization}

The starting point in the design of many popular on-policy algorithms is the following policy improvement lower bound, which was first developed by Kakade and Langford\citep{kakade2002approximately} and later refined by Schulman\citep{schulman2015trust}:

\textbf{\emph{Lemma 1.}} Consider a current policy $\pi_k$. For any future policy $\pi$, we have:
\begin{equation}\label{eq:3}
    \begin{aligned}
    J(\pi)-J(\pi_k)&\geq\frac{1}{1-\gamma}\mathbb{E}_{(s,a)\sim d^{\pi}_k}[\frac{\pi(a|s)}{\pi_k(a|s)}A^{\pi_k}(s,a)]-\\
  &\frac{2\gamma C^{\pi,\pi_k}}{(1-\gamma)^2}\mathbb{E}_{s\sim[TV(\pi,\pi_k)(s)] d^{\pi_k}} 
    \end{aligned}
\end{equation}

where $C^{\pi,\pi_k}=max_{s\in S}|\mathbb{E}_{a\sim \pi(\cdot|s)}[A^{\pi_k}(s,a)]|$ and $TV(\pi,\pi_k)(s)$ represents the total variation distance between the distributions $\pi(\cdot|s)$ and $\pi_{k}(\cdot|s)$ .

The first term of the lower bound in Lemma 1 is the surrogate objective, and the second term is the penalty term. Note that this approach can guarantee policy improvement at every step of the learning process by choosing the next policy $\pi_{k+1}$ to maximize this lower bound. Because the expectations in Lemma 1 depend on the current policy $\pi_{k}$, we can approximate this lower bound using samples generated by the current policy.

PPO\citep{schulman2017proximal}, which has become the default on-policy optimization algorithm due to its strong performance and simple implementation, is theoretically motivated by the policy improvement lower bound in Lemma 1.

Based on Lemma 1, the PPO defines the following surrogate objectives:
\begin{equation}\label{eq:4}
\begin{aligned}
\max {L^{CPI}}(\theta) &= \max{\hat{\mathbb{E}_t}}[{r_t}(\theta ){\hat A_t}],\\{r_t}(\theta ) &= \frac{{{\pi _\theta }({a_t}|{s_t})}}{{{\pi _{{\theta _{old}}}}({a_t}|{s_t})}}{\rm{ }}
\end{aligned}
\end{equation}
where $\pi_{\theta}$ represents the current policy, $\pi_{\theta_{old}}$ represents the old policy that generated the sample, and $\hat {A_t}$ estimates the value of the advantage function of action $a_t$ in state $s_t$. Equation 4 aims to maximize the value of the surrogate objective while constraining the distance between the new policy and the old policy. Moreover, PPO trims the above surrogate objective.
\begin{equation}\label{eq:5}
  L^{\text {CUIP }}(\theta)=\hat{\mathbb{E}}_{t}\left[\min \left(r_{t}(\theta) \hat{A}_{t}, \operatorname{clip}\left(r_{t}(\theta), 1-\varepsilon, 1+\varepsilon\right) \hat{A}_{t}\right]\right.
\end{equation}
\begin{equation}\label{eq:6}
 \operatorname{clip}(x, x_{MIN}, x_{MAX})=\left\{\begin{array}{cc}
x, & \text {if} x_{MIN} \leq x \leq x_{MAX} \\
x_{MIN}, & \text { if } x<x_{MIN} \\
x_{MAX}, & \text {if} x_{MAX}<x
\end{array}\right.
\end{equation}

\subsection{Importance Sampling}
For most practical applications of probabilistic models, exact inference is not feasible; some form of approximation must be used. Here, we discuss the use of a probability distribution $p(z)$ to find the expected value of some function $f(z)$, where the element $z$ may be a discrete variable, a continuous variable or a combination of both. In the case of a continuous variable we wish to calculate the following expectation: $\mathbb{E}[f]=\int f(z)p(z)dz$. In the case of a discrete variable the integral term is replaced by a summation term.

Importance sampling\citep{sutton2018reinforcement,braverman2021adversarial} provides a framework for approximating expectations, but does not provide a way to sample from the probability distribution $p(z)$. It is assumed that sampling directly from the distribution $p(z)$ is not possible, but for any given value of $z$, we can easily compute $p(z)$. Importance sampling is based on the use of the proposed distribution $q(z)$. It is relatively easy to sample from the proposed distribution, after which the expectation is expressed in the form of a finite sum of samples $\{z^{l}\}$ in $q(z)$, as shown in the following equation.

\begin{equation}\label{eq:7}
   \begin{aligned}
\mathbb{E}[f] &=\int f(z) p(z) d z \\
&=\int f(z) \frac{p(z)}{q(z)} q(z) d z \\
& \simeq \frac{1}{L} \sum_{l=1}^{L} \frac{p\left(z^{(l)}\right)}{q\left(z^{(l)}\right)} f\left(z^{(l)}\right)
\end{aligned} 
\end{equation}
where $r_{l}=\frac{p\left(z^{(l)}\right)}{q\left(z^{(l)}\right)}$ is referred to as the importance weight and corrects for the bias introduced by sampling from the wrong probability distribution.

Off-policy reinforcement learning algorithms \citep{mnih2015human} reuse early experience based on importance sampling, and the return is weighted according to the relative probability of the trajectory occurring in the target and behavior policy. Given a starting state $s_t$ , the probability of a subsequent state-action trajectory $a_t,s_{t+1},a_{t+1},...,s_T$ occurring under policy $\pi$ is:
\begin{equation}\label{eq:8}
    \begin{array}{l}
P\left\{a_{t}, s_{t+1}, a_{t+1}, \ldots, s_{T} \mid s_{t}, a_{t: T-1} \sim \pi\right\} \\
=\pi\left(a_{t} \mid s_{t}\right) p\left(s_{t+1} \mid s_{t}, a_{t}\right) \pi\left(a_{t+1} \mid s_{t+1}\right) \cdots p\left(s_{t} \mid s_{t-1}, a_{t-1}\right) \\
=\prod_{k=t}^{T-1} \pi\left(a_{k} \mid s_{k}\right) p\left(s_{k+1} \mid s_{k}, a_{k}\right)
\end{array}
\end{equation}
where $p$ denotes the environmental state shift probability function.

The importance sampling ratio of the target $\pi$ and behavior policy $b$ is:
\begin{equation}\label{eq:9}
    \begin{aligned}
    \rho_{t:T-1} &=\frac{\prod_{k=1}^{T-1} \pi\left(a_{k} \mid s_{k}\right) p\left(s_{k+1} \mid s_{k}, a_{k}\right)}{\prod_{k=t}^{T-1} b\left(a_{k} \mid s_{k}\right) p \left(s_{k+1} \mid s_{k}, a_{k}\right)}\\
    & =\prod_{k=t}^{T-1} \frac{\pi\left(a_{k} \mid s_{k}\right)}{b\left(a_{k} \mid s_{k}\right)}
    \end{aligned}
\end{equation}
From equation\ref {eq:9}, it is clear that the importance sampling ratio is related to only the policy and sample sequence data, and not to the dynamic characteristics of the environment.

\section{Trajectory Priority}
Prioritized trajectory replay (PTR) is based on the PER method of sampling trajectories according to importance. In the PER-DQN method, the priority of one-step experience is calculated using equation \ref {eq:1}. One-step experience TD-error calculations have small variance but excessive bias, whereas trajectories contain multistep experience and have less bias but larger variance in the estimate.

To better weigh the relationship between variance and bias, GAE\citep{schulman2015high} considers bias estimates from multistep experience and obtains an estimate of the advantage of the current action value:
\begin{equation}\label{eq:10}
    A_{t}=\delta_{t}+(\gamma \lambda) \delta_{t+1}+\cdots+(\gamma \lambda)^{T-t-1} \delta_{T-1}
\end{equation}

\begin{equation}\label{eq:11}
    \delta_{t}=r_{t}+\gamma V\left(s_{t+1}\right)-V\left(s_{t}\right)
\end{equation}
where $\gamma$ and $\lambda$ are hyperparameters

We use the GAE algorithm to calculate the bias in the advantage estimate for each experience in the trajectory, and equation \ref {eq:10} defines the bias in the advantage of one-step experiences. The trajectory $\tau_{i}=\left\{\left(s_{i}, a_{i}, r_{i+1}, s_{i+1}, d_{i+1}, p_{i}\right)\right\}$ contains multiple steps of experience, and the priority $p_{\tau}^{i}$ of the trajectory $\tau_i$ is related to the advantage bias of the same one-step experience. The priority of a trajectory is defined based on the GAE deviation of one-step experience, and two means of calculating the priority of a trajectory are defined.

1) Max trajectory priority: trajectory priority based on the maximum factor of the one-step experience GAE deviation. As with the PER method, experiences with larger deviations should receive more attention, regardless of the positive or negative value. The priority of a trajectory is obtained by applying the max operation to the priority of the one-step experiences contained in the trajectory, using the absolute value $|A_t|$ to denote the priority of the one-step experience.
\begin{equation}\label{eq:12}
    p_{\tau}^{i}=\max\{|A_j|\}_{\tau}^{i}
\end{equation}

2) Mean trajectory priority: trajectory prioritization based on one-step empirical GAE deviations of the mean factor. As with the max trajectory priority, the priority of the one-step experience is expressed using the absolute value $|A_t|$, but the priority of the trajectory is obtained by taking the mean operation on the priority of one-step experiences.
\begin{equation}\label{eq:13}
    {p}_{\tau}^{i}=\text{mean}\{|A_j|\}_{\tau}^{i}
\end{equation}

\begin{figure}[!ht]
\centering
\includegraphics[width=7cm,height=3.5cm]{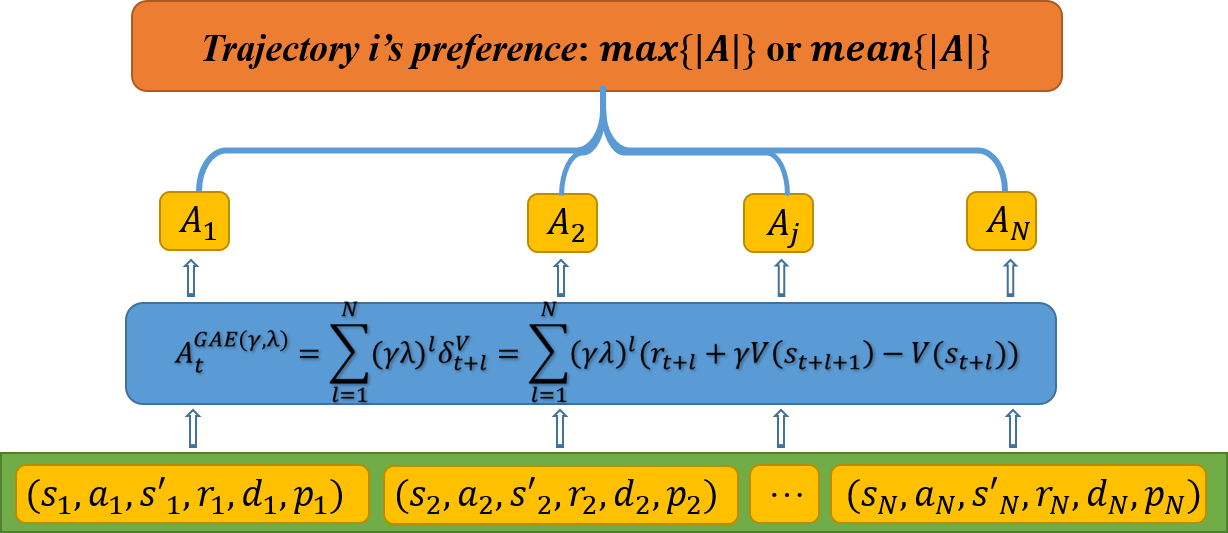}
\caption{Trajectory priority based on one-step experience GAE deviation. The one-step empirical advantage deviation for each experience in the trajectory is calculated using GAE and the appropriate value is obtained as the trajectory priority using the max and mean operators.}
\label{fig:my_label}
\end{figure}

In addition, a third trajectory priority is defined in this paper.

3) Reward trajectory priority. We define a trajectory priority based on normalized undiscounted cumulative reward. Since the length of a trajectory is finite, the return, that is, the undiscounted cumulative reward, does not lead to divergent results. To better measure the undiscounted cumulative reward, the undiscounted cumulative reward of the current trajectory are normalized based on the historical average and variance.
\begin{equation}\label{eq:14}
    \begin{aligned}
    P_{z}^{i} &=\left|\left(R^{i}-\bar{X}\right) / \sigma\right|\\
\bar{X}_{t} &=\bar{X}_{t-1}+\left(R_{t}-\bar{X}_{t-1}\right) /|\{\tau\}| \\
\sigma_{t}^{2} &=\sigma_{t-1}^{2}+\left(R_{t}-\bar{X}_{t-1}\right)\left(R_{t}-\bar{X}_{t}\right) 
  \end{aligned}
\end{equation}
where $R^i$ is the undiscounted cumulative reward for the current trajectory, $\overline{X_t}$ is the mean cumulative reward for all current collected historical trajectories, and $\sigma$ is the standard deviation.

\begin{figure}[!ht]
\centering
\includegraphics[width=7.15cm,height=2.72cm]{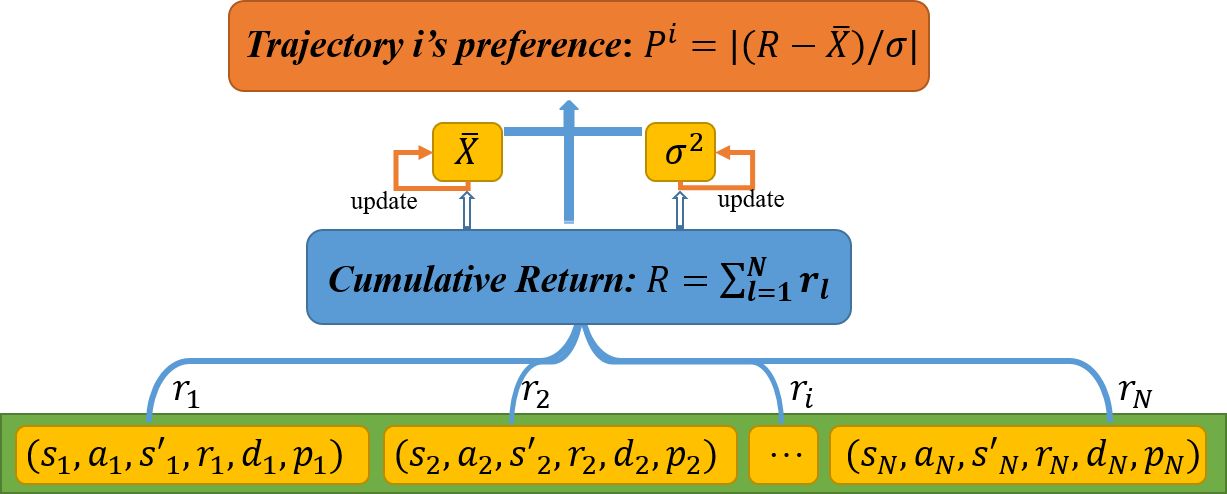}
\caption{Trajectory priority based on undiscounted cumulative reward. The normalized undiscounted cumulative reward for trajectory $i$ are calculated by obtaining the mean and variance of the returns for the current historical trajectory, updating the mean and variance of the cumulative reward based on the return of the current trajectory, and then obtaining the priority of the trajectory based on the new values. Note that the mean and variance of the return are updated only the first time the trajectory enters \emph{sumtree memory} in Fig. \ref{fig:arch}, after which only these two values are relied upon to calculate the priority without updating.}
\label{fig:rew}
\end{figure}

\section{PPO with Prioritized Trajectory Replay}
In this section, based on the trajectory priority mentioned in the previous section, we combine the trajectory priority with the PPO algorithm and propose the PPO with prioritized trajectory replay (PTR-PPO) algorithm with truncation importance weights. In this paper, we use a network with parameters $\theta$ shared by the actor $\pi_{\theta}(a|s)$ and critic $V_{\theta}(s)$, and the shared network outputs each action probability and the current state value at the last layer. We now provide the specific procedure for updating the network. 

\subsection{PTR-PPO Policy Evaluation}
In this paper, we update the state value function using a general policy evaluation approach, as in the following equation.

\begin{equation}\label{eq:15}
    J_{V}(\theta)=\mathop{\mathbb{E}}\limits_{(s,r,s',d,p)\in M}(r+V(s')-(1-d)V(s))^2
\end{equation}
where $(s,r,s',d,p)$ denotes the one-step experience in the trajectory and $M$ denotes the trajectory memory, namely, the current trajectory memory and the sampling trajectory memory. If the experience comes from the current trajectory memory, it is an on-policy learning approach and does not require a correction to the value estimate; conversely, if the experience comes from the sampling trajectory memory, the generation of the action $a$ comes from the old policy $\pi_{\theta_{old}}$, and the calculation of the loss according to Equation \ref {eq:15} is biased and must be learned using an off-policy learning approach, using the importance sampling ratio mentioned in Equation \ref {eq:9} for correction, calculated as:
\begin{equation}\label{eq:16}
    J_{V}(\theta)=\mathop{\mathbb{E}}\limits_{(s,r,s',d,p)\in M}\frac{\pi_{\theta}(a|s)}{\pi_{\theta_{old}}(a|s)}(r+V(s)-(1-d)V(s'))^2
\end{equation}
Since we record the probabilities $p$ at the time of the sampled action $a$, we do not need to store the parameters $\theta_{old}$ of the old network.

\subsection{PTR-PPO Policy Improvement}
As with policy evaluation, the experience in policy improvement also comes from two memories, the current trajectories and the sampling trajectories. For the policy improvement objective function in the current trajectories, a combination of Equation \ref {eq:4} and Equation \ref {eq:5} can be used, but the action advantage value $A_t$ is obtained using the GAE algorithm calculation described in Equation \ref{eq:10}.

Action values from the sampling trajectories, must be corrected them using importance sampling when calculating the action advantage values using the GAE method, as follows:
\begin{equation}\label{eq:17}
    A_{t}=\prod_{k=t}^{T-1} \frac{\pi\left(a_{k} \mid s_{k}\right)}{b\left(a_{k} \mid s_{k}\right)}[\delta_{t}+(\gamma \lambda) \delta_{t+1}+\cdots+(\gamma \lambda)^{T-t-1} \delta_{T-1}]
\end{equation}

In this paper, multiple environments are used to generate data simultaneously and each environment is simulated cyclically, i.e., a new simulation is started immediately after the end of the previous. With this setup, the done state will appear at an arbitrary position when the trajectory is stored according to a fixed step size. The importance sampling ratio at the end position should not be cumulatively multiplied by the subsequent ratios. Thus, Equation \ref{eq:11} and Equation \ref{eq:17} must be improved to adapt it to any trajectory:
\begin{equation}\label{eq:18}
    {A}_{t}^{done} =\delta_t+(1-d){A}_{t+1}^{done}, \delta_t=r+V\left(s_{t+1}\right)-V\left(s_{t}\right))
\end{equation}
\begin{equation}\label{eq:19}
    \rho_{t:T-1}^{done}  =[d+(1-d)\prod_{k=t+1}^{T-1} \frac{\pi\left(a_{k} \mid s_{k}\right)}{b\left(a_{k} \mid s_{k}\right)}] \frac{\pi(a_t|s_t)}{b(a_t|s_t)}
\end{equation}

When there is no done state in the trajectory or the done state is positioned very far back, the margins of the importance weights in Equation \ref{eq:19} become large, which causes instability in learning.
To prevent high variance, we propose truncation importance weights to constrain the bounds of the importance weights. As mentioned in ACER, we split the obtained importance weights into two terms: 
\begin{equation}\label{eq:20}
    \rho_{t:T-1}^{marg}=\text{min}({1-\epsilon^{marg},\rho_{t:T-1}^{done}})+[\frac{\rho_{t:T-1}^{done}-(1-\epsilon^{marg})}{\rho_{t:T-1}^{done}}]_+
\end{equation}
where $[x]_+=x$ if $x>0$ and otherwise $[x]_+=0$. The first term of Equation \ref{eq:20} constrains the upper bound on the importance weights, and the second term ensures that the weight estimate is unbiased, and adjust a large value to 1 when $\rho_{t:T-1}^{done}\gg (1-\epsilon^{marg})$. Therefore, the advantage value of GAE under the off-policy is: 
\begin{equation}\label{eq:21}
    {A}_{t}^{marg}=\rho_{t:T-1}^{marg}{A}_{t}^{done}
\end{equation}

Thus, the PTR-PPO policy improvement objective is defined by the following equation.
\begin{equation}\label{eq:22}
    J_{\pi}(\theta)=\mathbb{E}_t[min(r_t(\theta){A}_{t}^{marg},clip(r_t(\theta),1-\epsilon,1+\epsilon)){A}_{t}^{marg}]
\end{equation}
where $r_t(\theta)$ is defined in Equation \ref{eq:4}. In the policy gradient class of deep reinforcement learning, a regular term policy entropy loss $\mathcal{H}(a,\theta)$ is typically added to increase the exploratory nature of the policy, and by observing the loss value of the policy entropy, it is possible to determine the degree of improvement of the policy. Note that, according to the principles of the PPO algorithm, we must record the parameters of the old network $\theta_{old}$ to ensure stable improvement in the performance of the policy.

\begin{algorithm}[htpb]
\caption{PTR-PPO algorithmn} 
\label{al:1}
\hspace*{0.02in} {\bf Input:} 
target network parameters $\theta$, old network parameters $\theta_{old}$, max steps $MAX\_EPI$, rollout legth $len$, Priority memory $D$, Number of off-policy iteration $N$,  $\epsilon$ in PPO, $\epsilon^{marg}$ in importance weigths\\
\hspace*{0.02in} {\bf Output:} 
Optimal target network parameters $\theta^{*}$
\begin{algorithmic}[1]
\State Initialize network parameters $\theta$,$\theta_{old} \leftarrow \theta$
\State Initialize the multiple environment, get the state $s_0$
\State some description 
\For{$epi=1,...,i,...,MAX\_EPI$} 
    \State Initialize temp memory $\mathcal{T}=[]$
    \State Initialize learning rate $lr$,and parameter $\beta$
    \For{$t=1,...,len$}
        \State Select action $a\sim\pi_{\theta}(a_t|s_t)$
        \State Observe reward $r_t$, next state $s_{t+1}$, done $d_t$, action probility $p_t$
        \State Store transition tuple $(s_t,a_t,r_t,s_{t+1},d_t,p_t)$ in buffer $\mathcal{T}$
        \If{$d_t$ is True}
            \State Reset the environment, get the state $s$
            \State $s_{t+1}\leftarrow s$
        \EndIf
    \EndFor
    \State Backup network parameters $\theta_{old} \leftarrow \theta$
    \State Use PPO to update the target network:  \quad \quad \quad  \quad \quad $\theta \leftarrow\theta-lr\nabla_{\theta}[(J_V(\theta)+L^{CLIP}(\theta)+\beta\mathcal{H}(\cdot,\theta)]$
    \State Calculate the priority $p_{\tau_{j}}$of every trajectory $\tau_j$ in memory $\mathcal{T}$
    \State Store every trajectory [$(p_{\tau_{j}},\tau_j)$] in memory $D$
    \For{$train\_num=1,..,N$}
        \State Sample M trajectories $[(\tau_i,index_i,p_{\tau_{j}})]$ from $D$
        \State Backup network parameters $\theta_{old} \leftarrow \theta$ 
        \State Use PTR-PPO to update the target network:  \quad \quad \quad          $\theta \leftarrow \theta-lr\nabla_{\theta}J(\theta)$
        \State Update the priority $p_{\tau_{i}}$of every trajectory $\tau_i$ in $[(\tau_i,index_i,p_{\tau_{j}})]$ to memory $D$
    \EndFor
\EndFor
\end{algorithmic}
\end{algorithm}
\subsection{PTR-PPO Algorithm Pseudocode}
Finally, according to Equation \ref{eq:16} and Equation \ref{eq:22},the loss of parameter updates for shared networks is given by the following equation.
\begin{equation}\label{eq:23}
    J(\theta)=J_{V}(\theta)+J_{\pi}(\theta)+\beta\mathcal{H}(\cdot,\theta)
\end{equation}

The PTR-PPO algorithm is detailed in Algorithm \ref{al:1};  Fig.\ref{fig:arch} shows the architecture of algorithm.

\section{Experimental Verification}

\subsection{Benchmarks}
\begin{figure}[!ht]
\centering
\includegraphics[width=8cm,height=8.33cm]{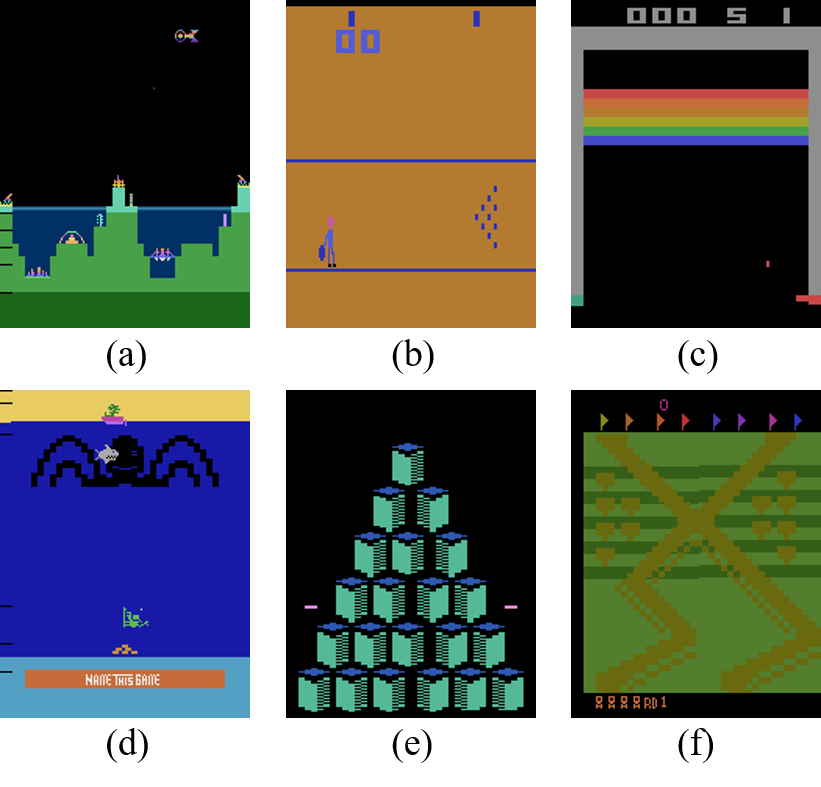}
\caption{Tasks.(a) Atlantis-v0: $(s\times a)\in \mathbb{R}^{28224}\times \mathbb{R}^4$. (b) Bowling-v0: $(s\times a)\in \mathbb{R}^{28224}\times \mathbb{R}^6$. (c) Breakout-v0: $(s\times a)\in \mathbb{R}^{28224}\times \mathbb{R}^4$. (d) NameThisGame-v0: $(s\times a)\in \mathbb{R}^{28224}\times \mathbb{R}^6$. (e) Qbert-v0: $(s\times a)\in \mathbb{R}^{28224}\times \mathbb{R}^6$. (f) UpNDown-v0: $(s\times a)\in \mathbb{R}^{28224}\times \mathbb{R}^6$}
\label{fig:game}
\end{figure}

To evaluate our algorithm, we tested its performance on a set of Atari games with discrete action spaces through the OpenAi Gym environment interface, without modifications to the environment. Fig. \ref{fig:game} illustrates the benchmark tasks used in this paper.

\subsection{Baselines}
We compare our proposed algorithm with the PPO and ACER methods, both mainstream reinforcement learning algorithms in the policy gradient class that have been validated and applied in many challenging domains. Using these algorithms as baseline algorithms allows us to effectively judge the performance of our proposed algorithms.
All compared algorithms interacted with the environment using the experience generation process shown in Fig. \ref{fig:arch}, containing 1 learner and 4 actors; in addition, they all had the same network architecture. The network consists of a convolutional layer with 32 8x8 filters with stride 4 followed by another convolutional layer with 64 4x4 filters with stride 2, a final convolutional layer with 64 3x3 filters with stride 1, and a fully connected layer of size 512. Each of the hidden layers is followed by a nonlinear rectifier. The network outputs a softmax policy and V values. The actor and critic networks share a common set of parameters.

\begin{table}[!ht]
    \centering
    \caption{Hyperparameters setting}
    \begin{tabular}{ccccccc}
        \hline  
        Hyperparameters& $\gamma$ & $\lambda$ & $lr$ & $\epsilon$ & $\epsilon^{marg}$ &$\beta$\\
        \hline  
        Value&0.99&0.95&1e-4&0.1&0.2&0.001\\
        \hline 
    \end{tabular}
    \label{tab:1}
\end{table}

The hyperparameter settings for the training process are shown in Table \ref{tab:1}.

\subsection{Results}

\subsubsection{Performance}
\begin{figure*}[htpb]
\centering
\subfigure[Altantis-v0]{\label{Altantis-v0}{\includegraphics[width=4.5cm,height=4cm]{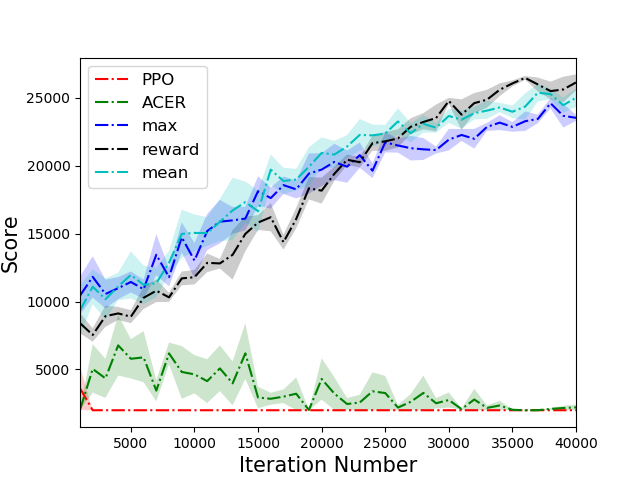}}}\hspace{1mm}
\subfigure[Bowling-v0]{\label{bowling-v0}{\includegraphics[width=4.5cm,height=4cm]{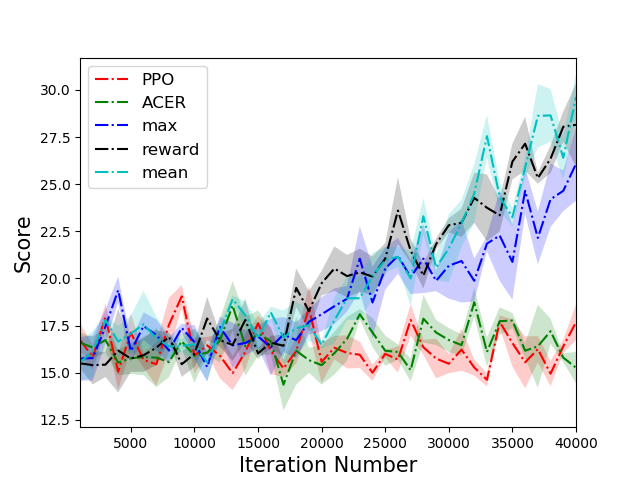}}}
\subfigure[Breakout-v0]{\label{Breakout-v0}{\includegraphics[width=4.5cm,height=4cm]{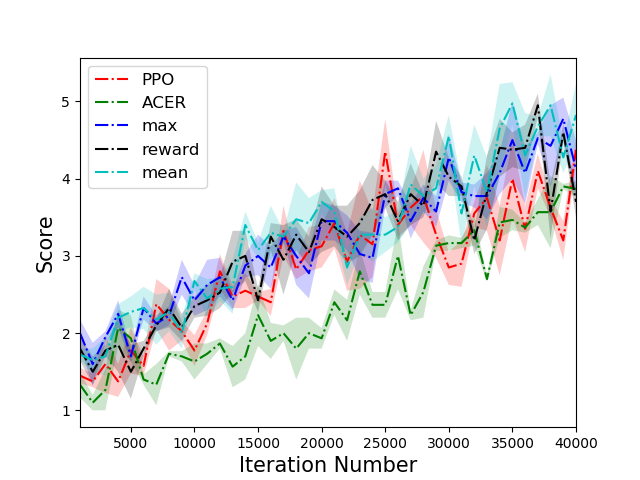}}}
\subfigure[NameThisGame-v0]{\label{NameThisGame}{\includegraphics[width=4.5cm,height=4cm]{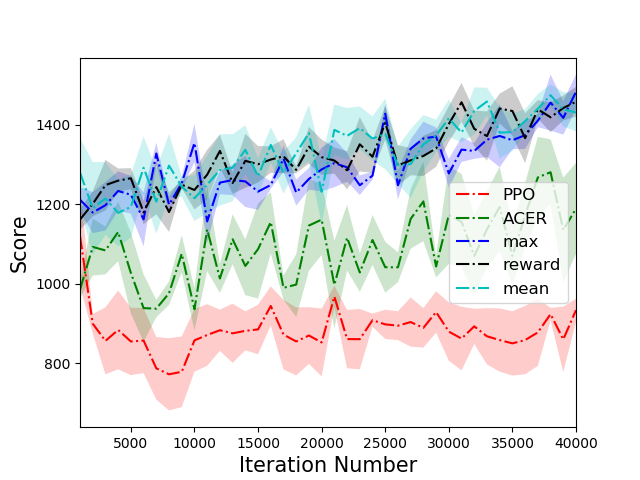}}}\hspace{1mm}
\subfigure[Qbert-v0]{\label{qbert-v0}{\includegraphics[width=4.5cm,height=4cm]{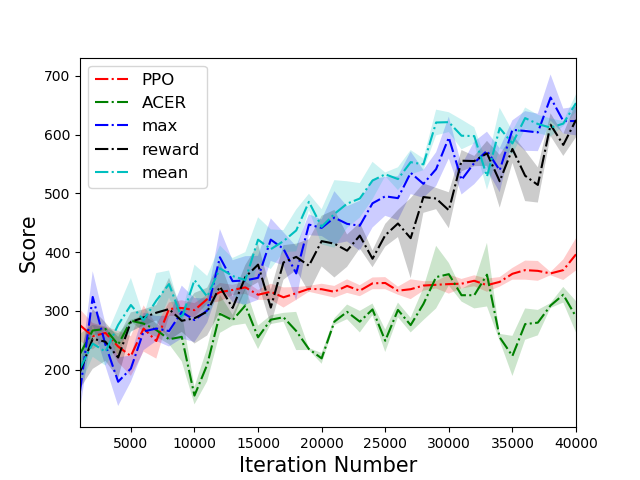}}}
\subfigure[UpNDown-v0]{\label{upndown-v0}{\includegraphics[width=4.5cm,height=4cm]{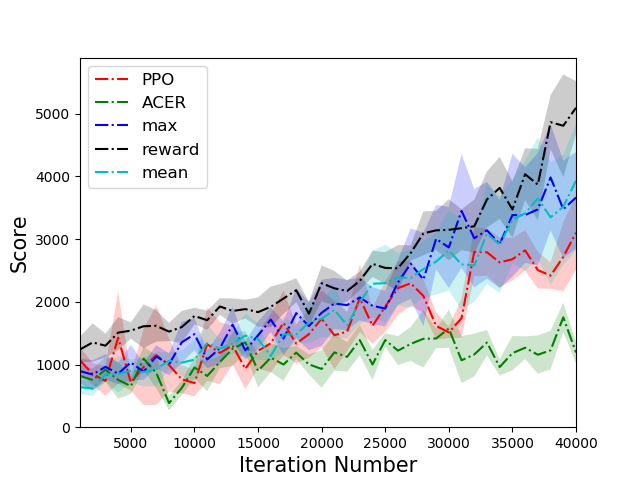}}}
\caption{Comparison of scores of different algorithms on 6 tasks. Red indicates the PPO algorithm, green indicates the ACER algorithm, dark blue indicates the PTR-PPO algorithm for max trajectory priority, black indicates the PTR-PPO algorithm for reward trajectory priority, and bright blue indicates the PTR-PPO algorithm for mean trajectory priority. The PTR-PPO algorithm significantly outperforms the PPO and ACER algorithms in five environments. In the Breakout-v0 environment, these algorithms score similarly, but PTR-PPO still has a small score advantage.}
\label{fig:ori_pic}
\end{figure*}

:
We conducted five training runs for each algorithm using random seeds, each lasting 40,000 steps, for a total of 160,000 experiences. Each execution of 1000 steps was followed by one model test, using 10 environments per test, and the mean of the test results was used as the final value for this test. The learning curve for each algorithm is shown in Fig. \ref{fig:ori_pic} and the results are shown in Table \ref{tab:2}. The experimental results indicate that the proposed PTR-PPO algorithm outperforms or matches the baseline algorithm in all tasks. Compared to the mainstream PPO and ACER algorithms, our algorithm is substantially ahead in all five tasks, including Altantis-v0, Bowling-v0, NameThisGame-v0, Qbert-v0 and UpNdown-v0. In the task Breakout-v0, our algorithm has similar results to these algorithms, but still has a significant advantage. This result indicates that our algorithms have exceeded the state of the art in these benchmarks.
\begin{table*}[!ht]
    \centering
    \caption{AVERAGE FINAL RETURN. MAXIMUM VALUE FOR EACH TASK IS BOLDED. CORRESPONDS TO A SINGLE STANDARD DEVIATION OVER 5 RUNS.}
    \resizebox{\textwidth}{12mm}{
    \begin{tabular}{cccccc}
        \hline  
        Tasks& PTR-PPO with max & PTR-PPO with mean & PTR-PPO with reward & PPO& ACER\\
        \hline  
        Altantis-v0&$23512\pm380$&$24990\pm1555$&$\mathbf{26117\pm1122}$&2000$\pm$0&2210$\pm$415\\
        Bowling-v0&$26.0\pm4.2$&$\mathbf{29.6\pm2.0}$&$28.2\pm3.8$&$15.3\pm1.2$&$17.6\pm1.7$\\
        Breakout-v0&$4.1\pm0.7$&$\mathbf{4.8\pm0.7}$&$3.7\pm0.4$&$3.9\pm0.2$&$4.4\pm0.4$\\
        NameThisGame-v0&$\mathbf{1482\pm84}$&$1430\pm81$&$1458\pm76$&$1189\pm225$&$933\pm68$\\
        Qbert-v0&$623\pm47$&$\mathbf{654\pm29}$&$625\pm51$&$290\pm54$&$396\pm56$\\
        UpNdown-v0&$3663\pm1500$&$3931.5\pm2025$&$\mathbf{5093\pm850}$&$1193\pm38$&$3101\pm1053$\\
        \hline 
    \end{tabular}
    \label{tab:2}}
\end{table*}

\subsubsection{Time Efficiency}
\begin{figure}[!ht]
\centering
\includegraphics[width=7cm,height=6cm]{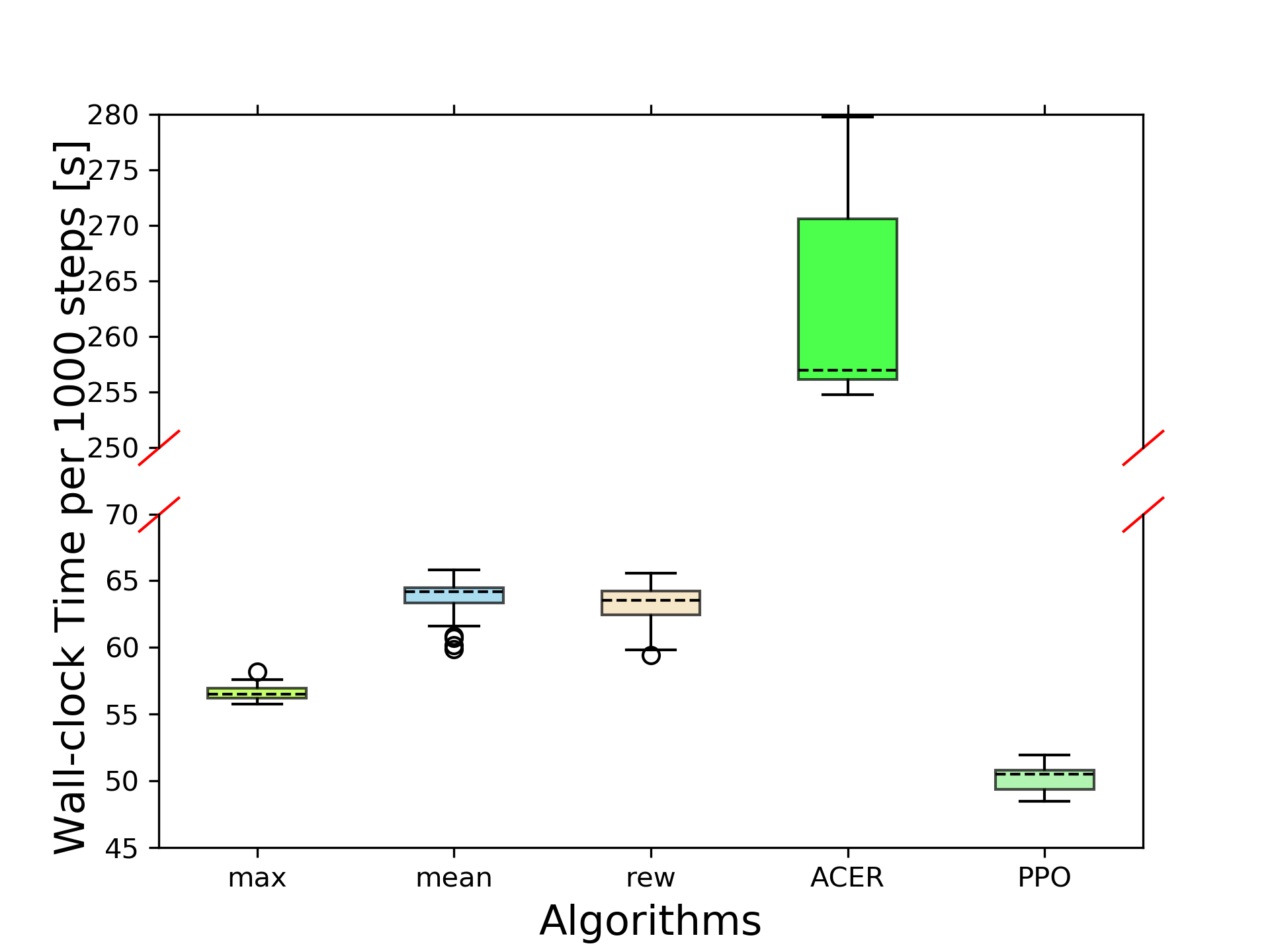}
\caption{Algorithm comparison in terms of time efficiency on the Altantis-v0 benchmark. Each boxplot is drawn based on the execution of 1000 steps, including approximately 62 improvements. All evaluations were performed on server with a 2.5 GHz 48 core Intel Xeon Platinum 8163 CPU.}
\label{fig:time}
\end{figure}

Fig. \ref{fig:time} compares the time efficiency of
different algorithms. The average wall-clock time consumption per 1000 steps of PTR-PPO is comparable to that of PPO and much lower than that of ACER. This is because the PTR-PPO algorithm maintains the same training times as the PPO algorithm, such as 10 training repetitions for the current trajectories in PPO, while PTR-PPO trains the current trajectories twice and samples 8 priority memory trajectories for training. PTR-PPO takes slightly longer than PPO, mainly for data dumping and format conversion. The ACER algorithm requires the use of a serial approach for one-step empirical computation, which takes longer to process the same length of trajectory.

\subsection{Ablation Studies}
Two hyperparameters are included in the PTR-PPO algorithm: 1) the priority memory size, which determines the replay history of the trajectory, with the empirical memory dropping to 0, the algorithm becomes a general PPO; 2) the trajectory rollout length, which indicates how many empirical steps are used for rollout. The longer the trajectory is, the more egalitarian the preference calculation for the trajectory, degrading to uniform sampling, and an overly short trajectory length will increase the bias of the GAE estimate.

\begin{figure*}[htpb]
\subfigure[Max trajectory priority ]{\label{Max_memory}{\includegraphics[width=5.5cm,height=4cm]{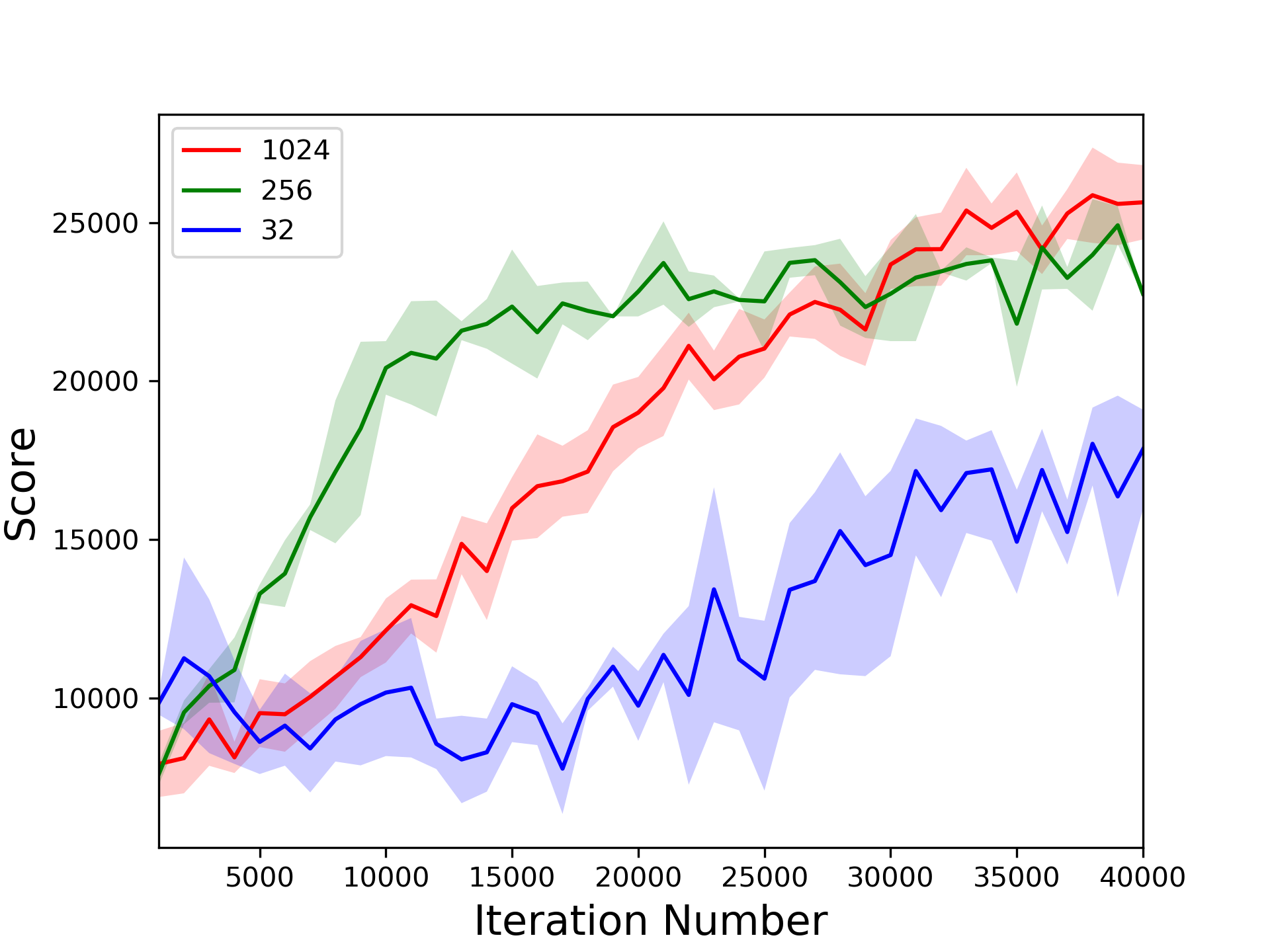}}}\hspace{1mm}
\subfigure[Mean trajectory priority ]{\label{Mean_memory}{\includegraphics[width=5.5cm,height=4cm]{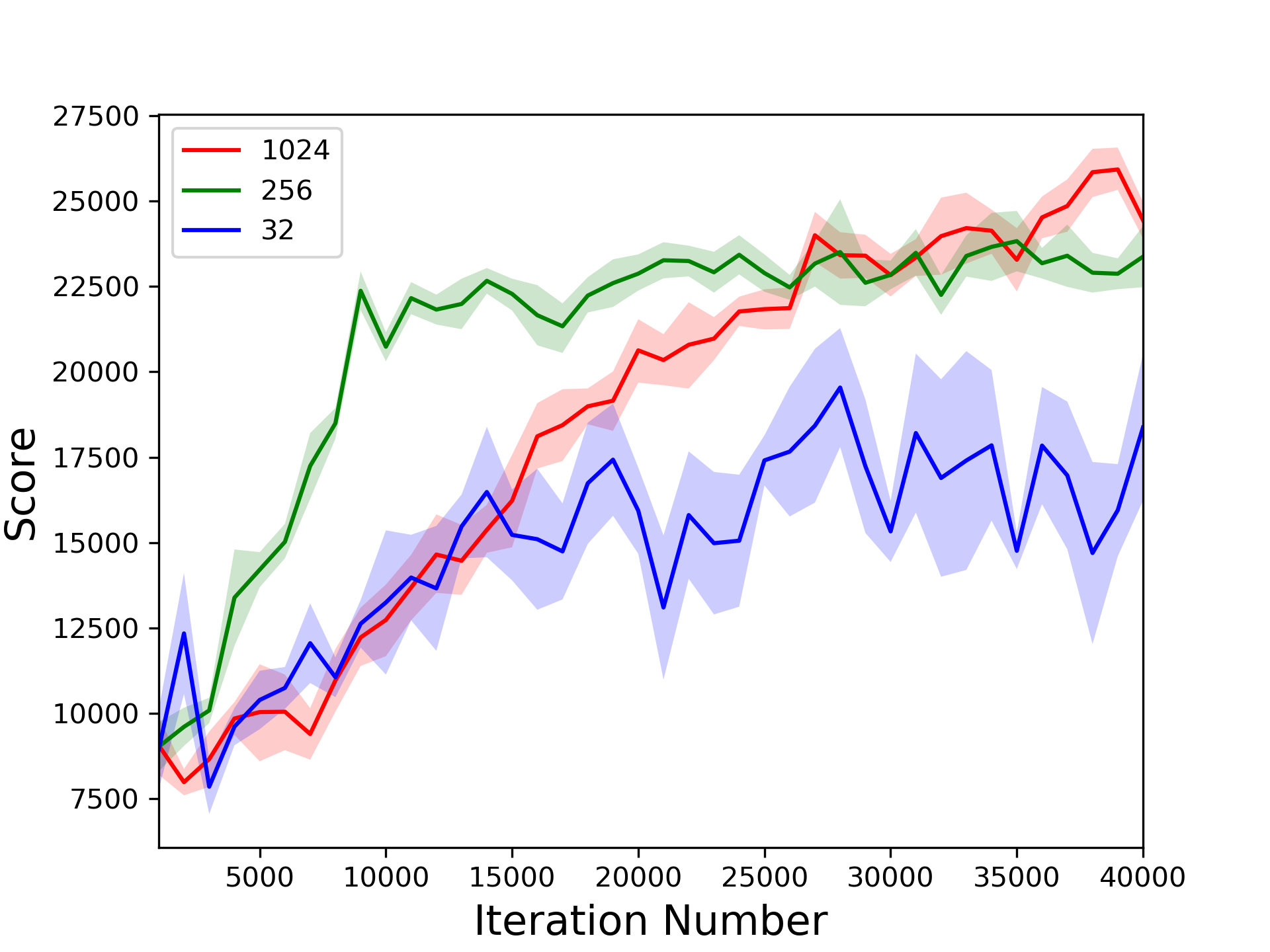}}}
\subfigure[Reward trajectory priority]{\label{Rew_memory}{\includegraphics[width=5.5cm,height=4cm]{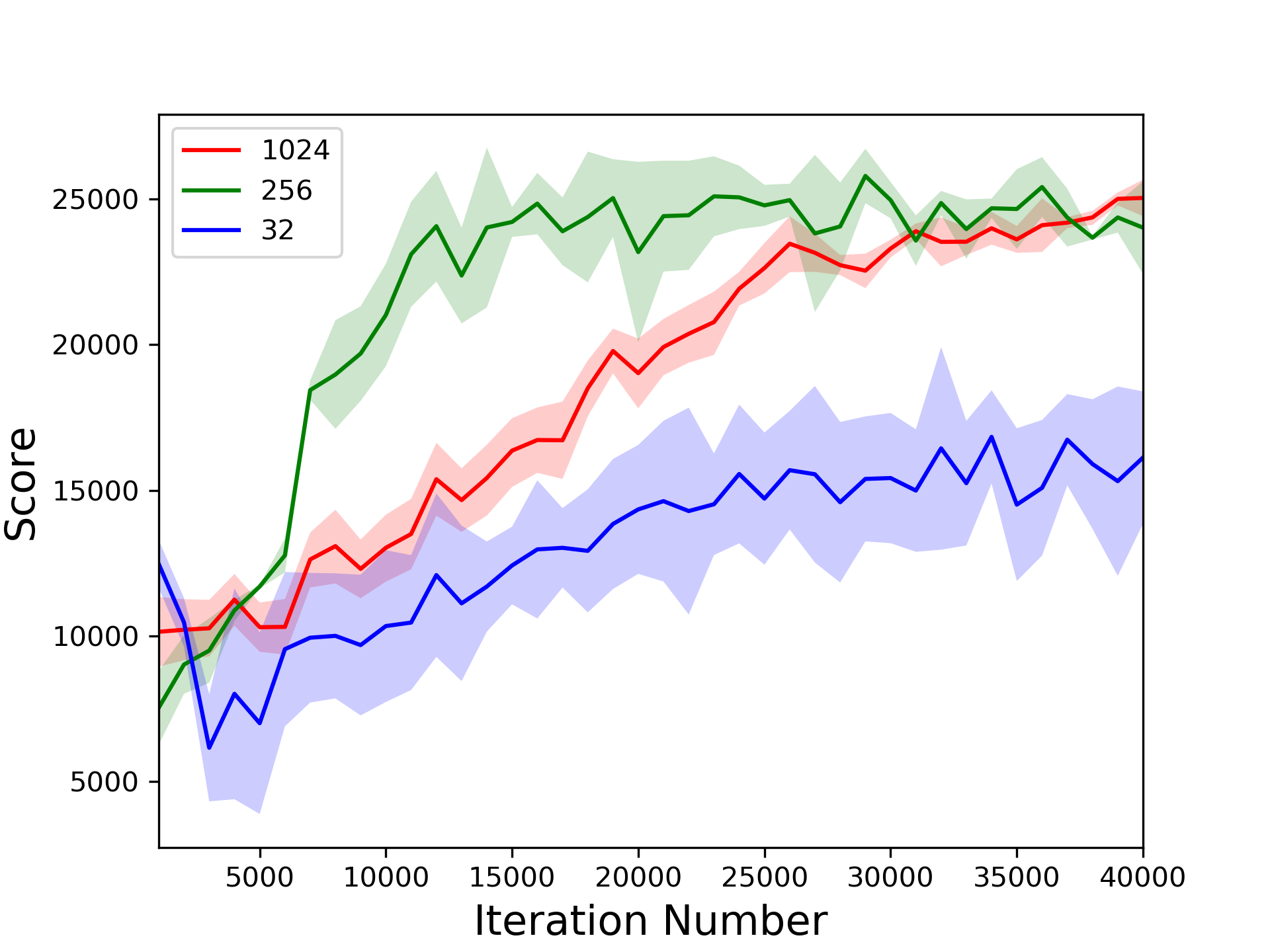}}}
\caption{Priority memory size comparisons. The red line indicates a memory size of 1024, the green line indicates a memory size of 256, and the blue line indicates a memory size of 32. (a) Comparison graph for the max trajectory priority, (b) comparison graph for the mean trajectory priority, and (c) comparison graph for the reward trajectory priority}
\label{fig:memory compare}
\centering
\subfigure[memory size of max priority= 1024 ]{\label{max_memory1024}{\includegraphics[width=4.5cm,height=4cm]{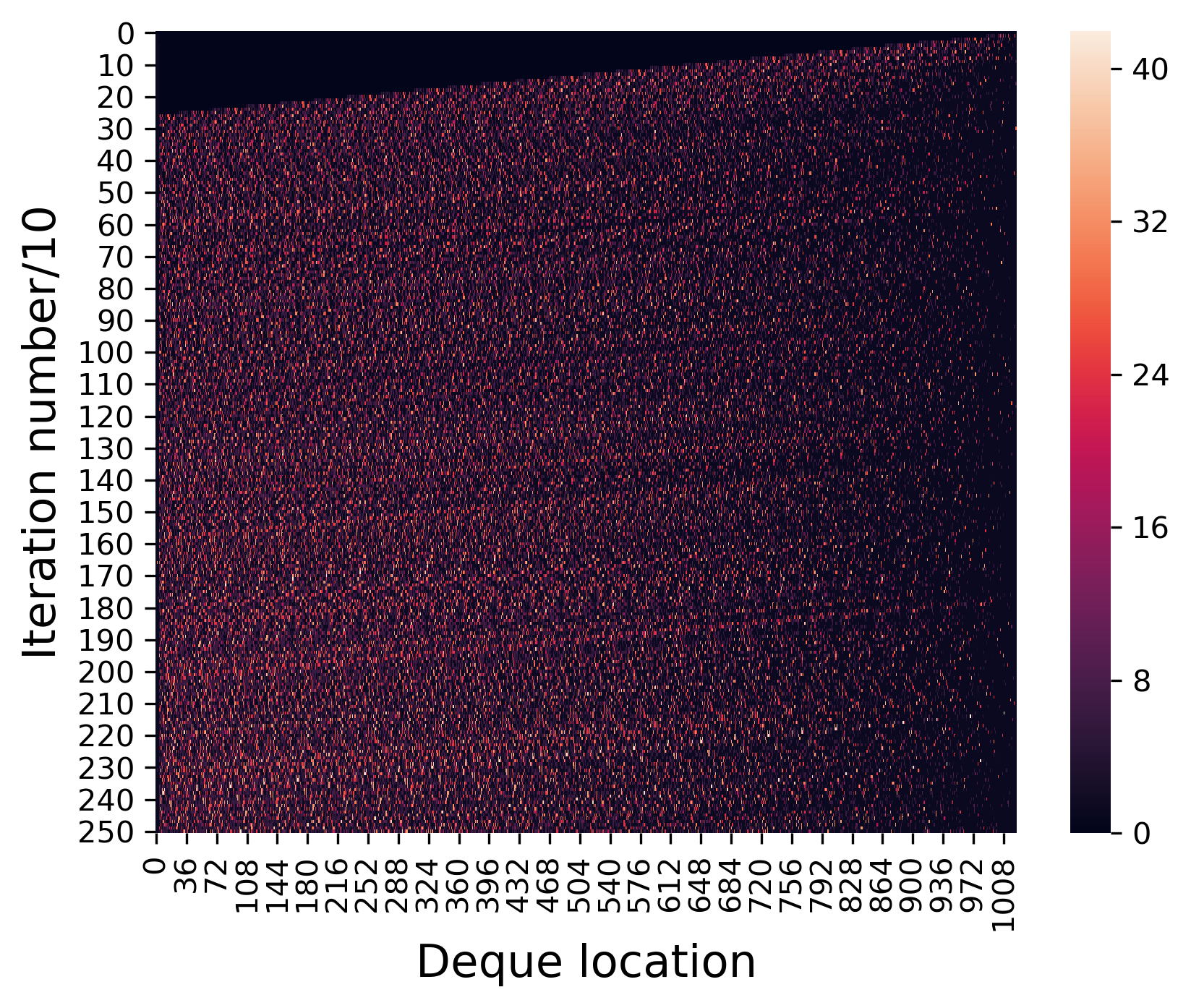}}}\hspace{1mm}
\subfigure[memory size of max priority= 256 ]{\label{max_memory256}{\includegraphics[width=4.5cm,height=4cm]{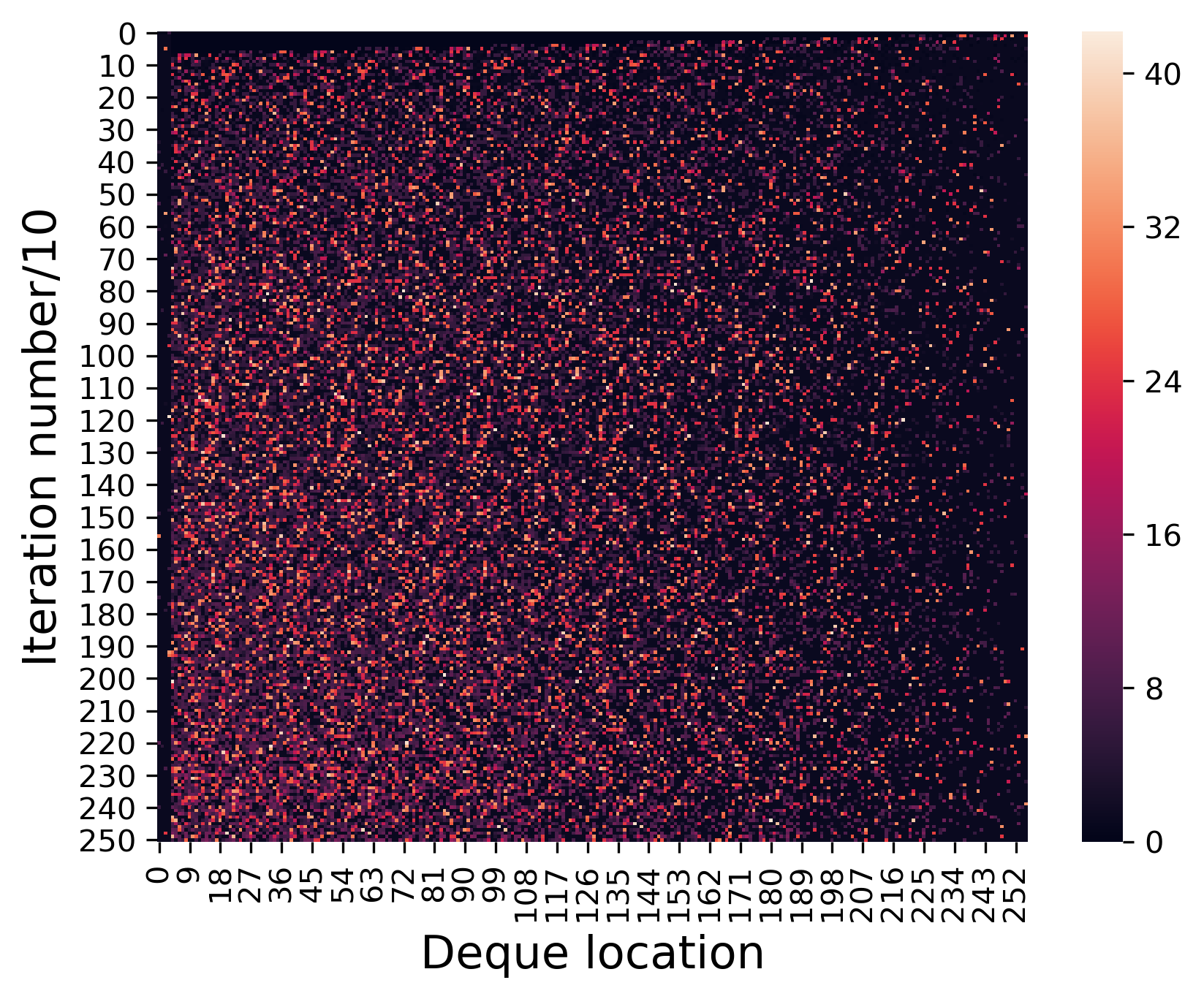}}}
\subfigure[memory size of max priority= 32]{\label{max_memory32}{\includegraphics[width=4.5cm,height=4cm]{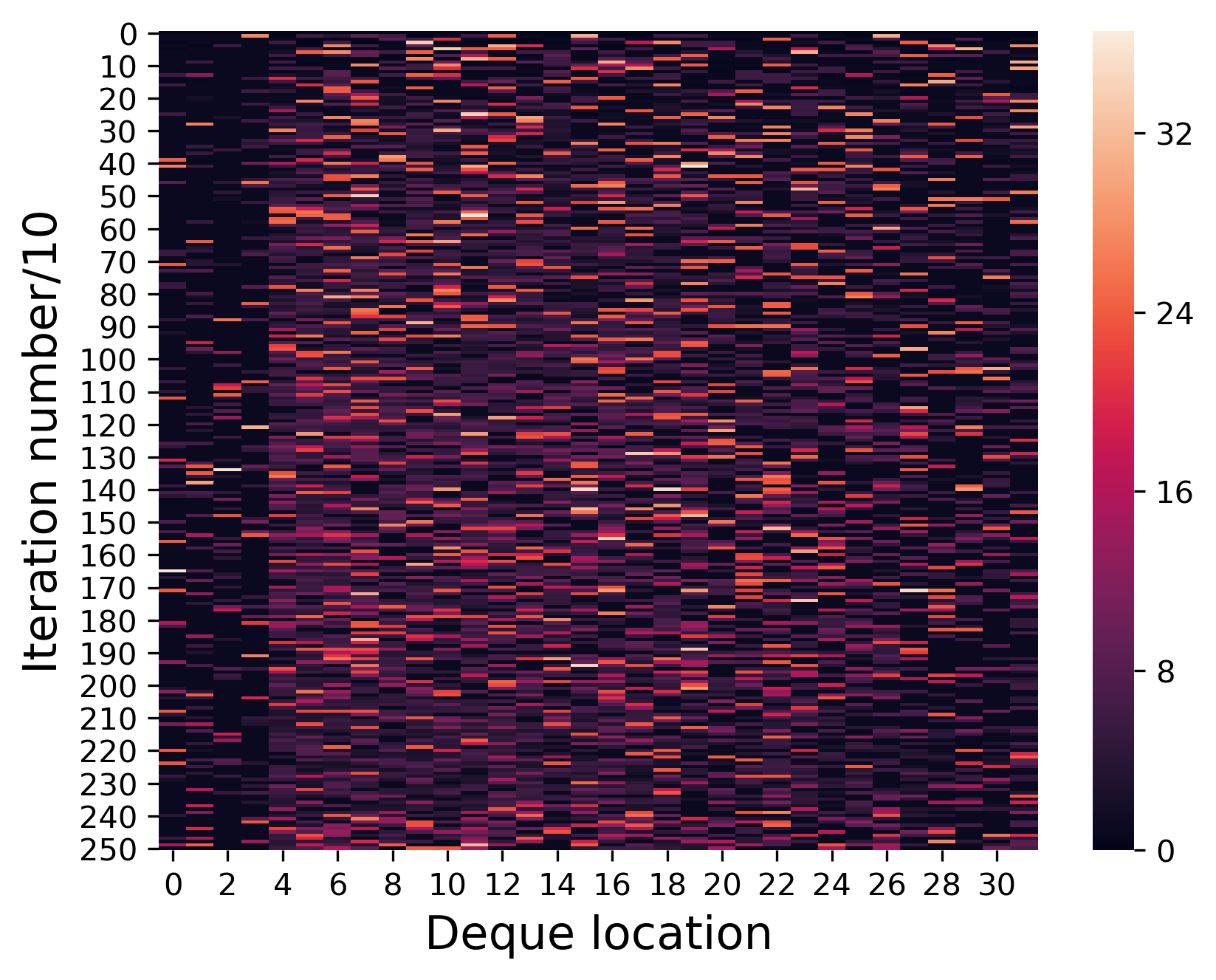}}}

\subfigure[memory size of mean priority= 1024 ]{\label{mean_memory1024}{\includegraphics[width=4.5cm,height=4cm]{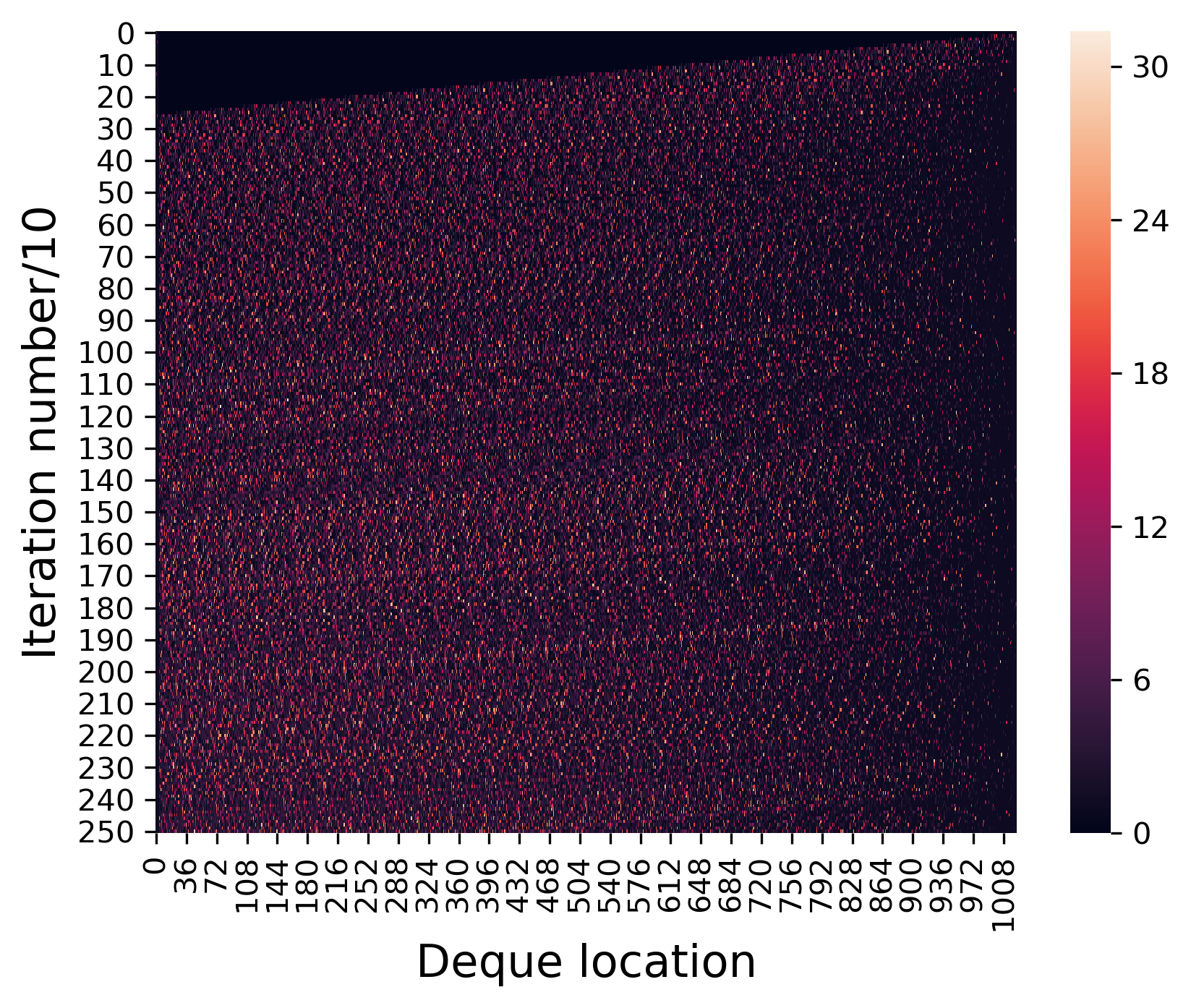}}}\hspace{1mm}
\subfigure[memory size of mean priority= 256 ]{\label{mean_memory256}{\includegraphics[width=4.5cm,height=4cm]{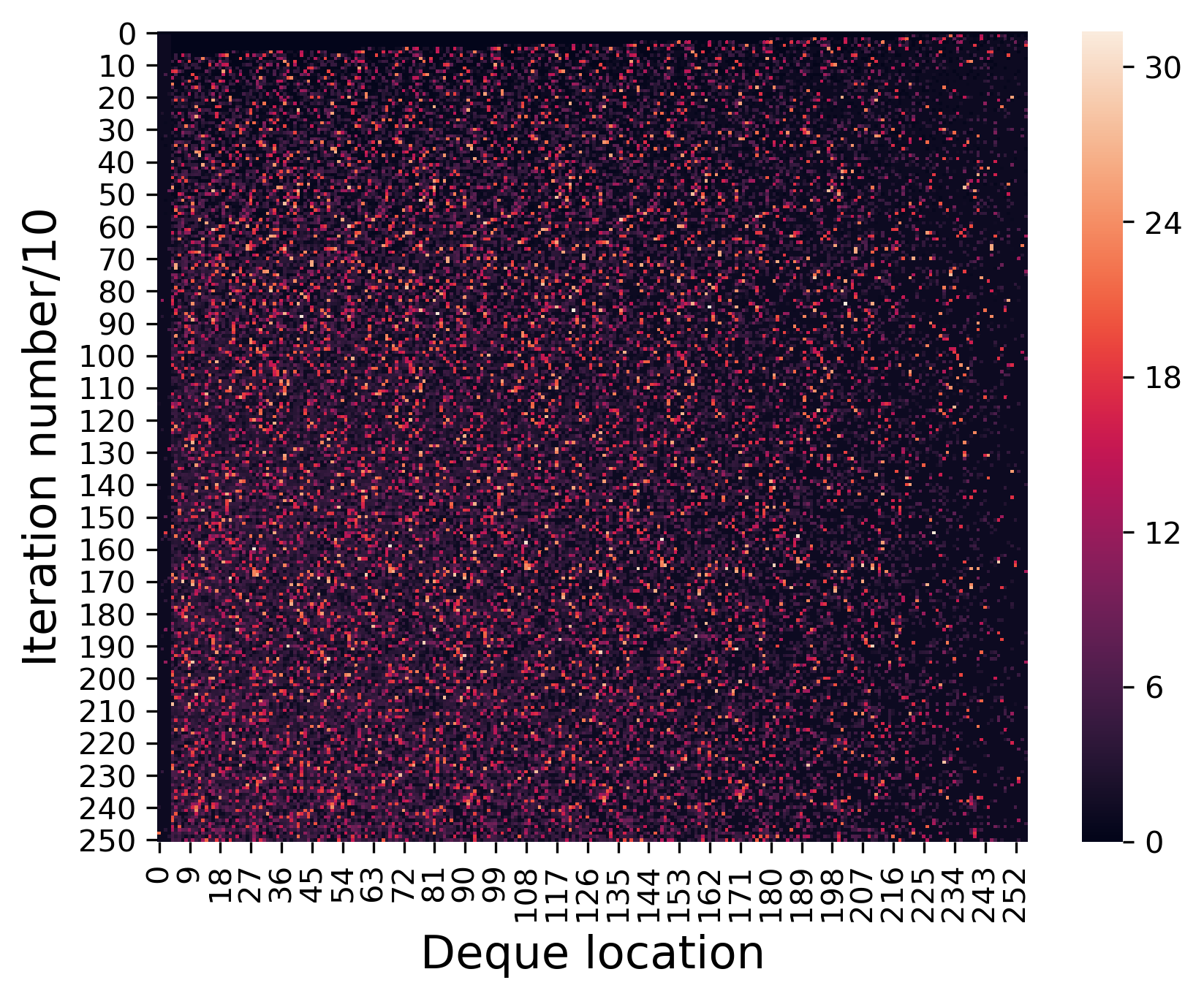}}}
\subfigure[memory size of mean priority= 32]{\label{mean_memory32}{\includegraphics[width=4.5cm,height=4cm]{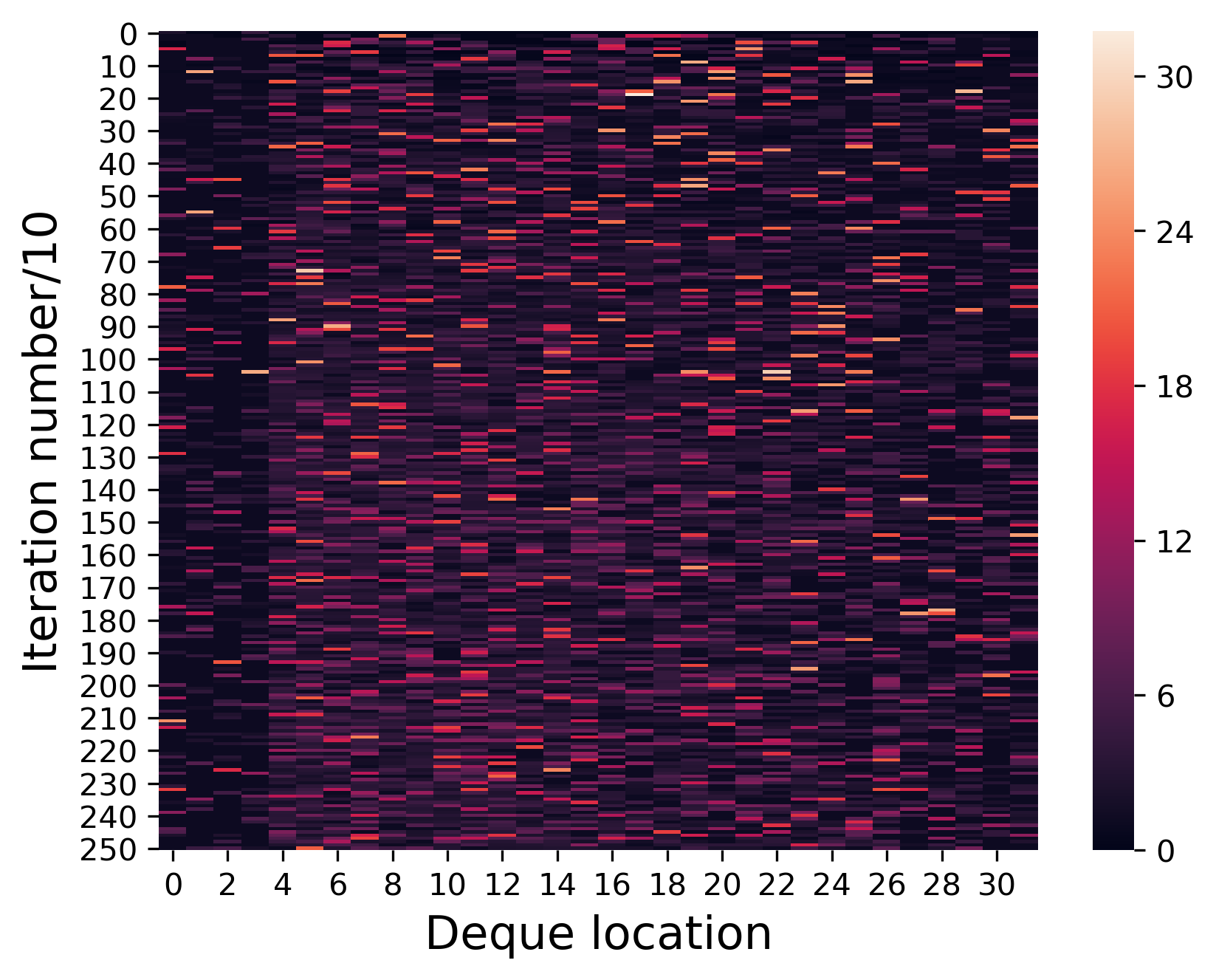}}}
\centering
\subfigure[memory size of reward priority= 1024 ]{\label{rew_memory1024}{\includegraphics[width=4.5cm,height=4cm]{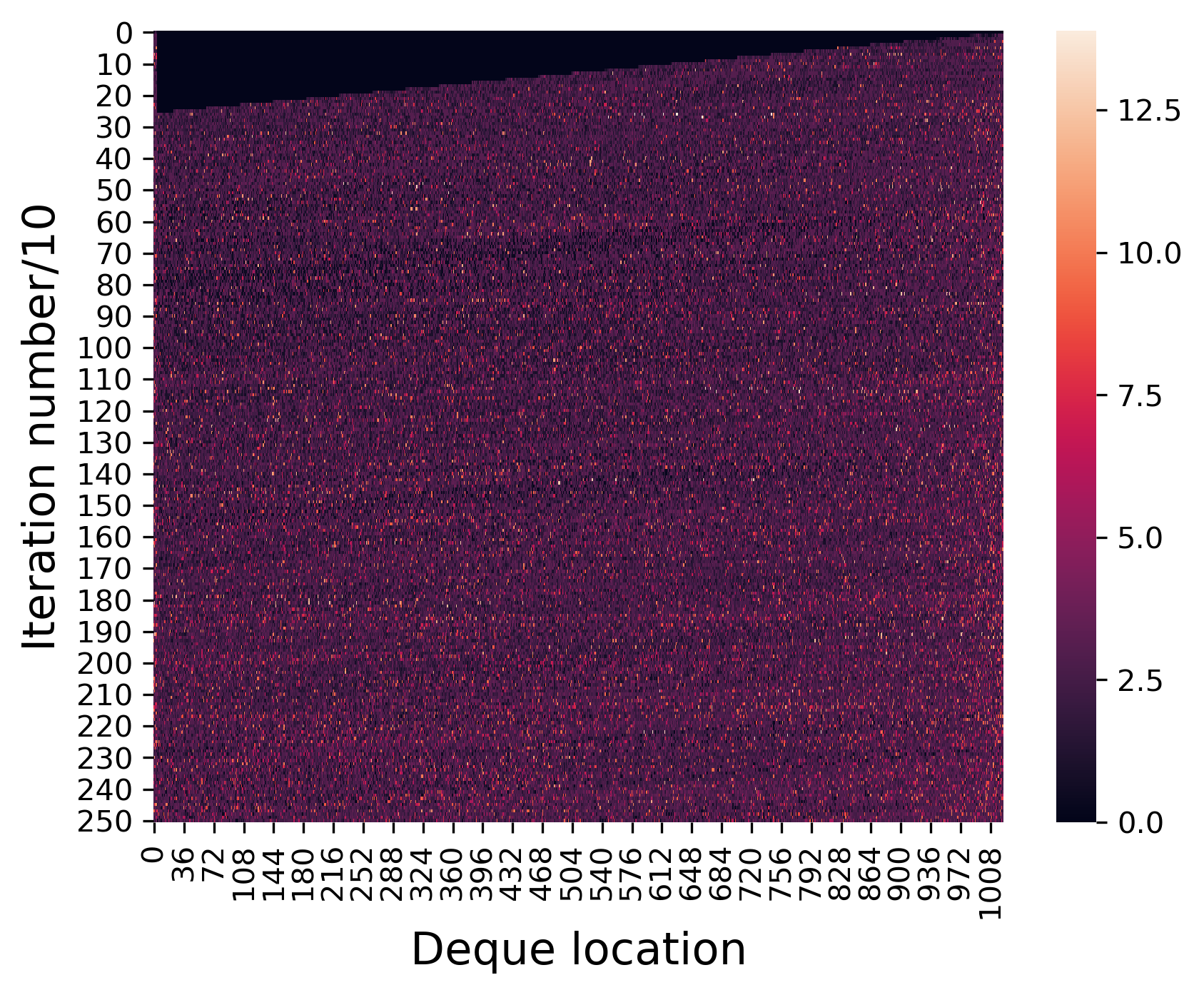}}}\hspace{1mm}
\subfigure[memory size of reward priority= 256 ]{\label{rew_memory256}{\includegraphics[width=4.5cm,height=4cm]{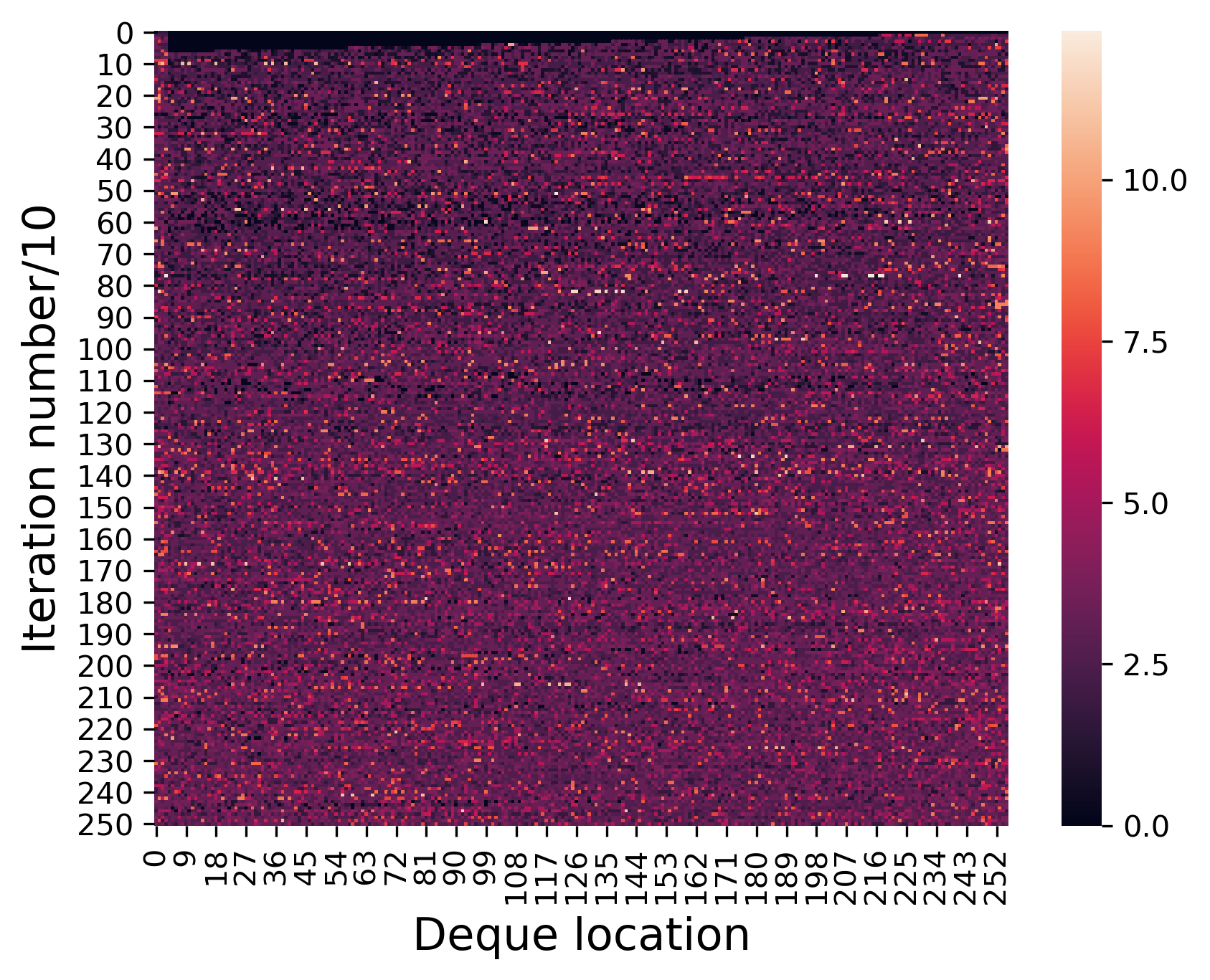}}}
\subfigure[memory size of reward priority= 32]{\label{rew_memory32}{\includegraphics[width=4.5cm,height=4cm]{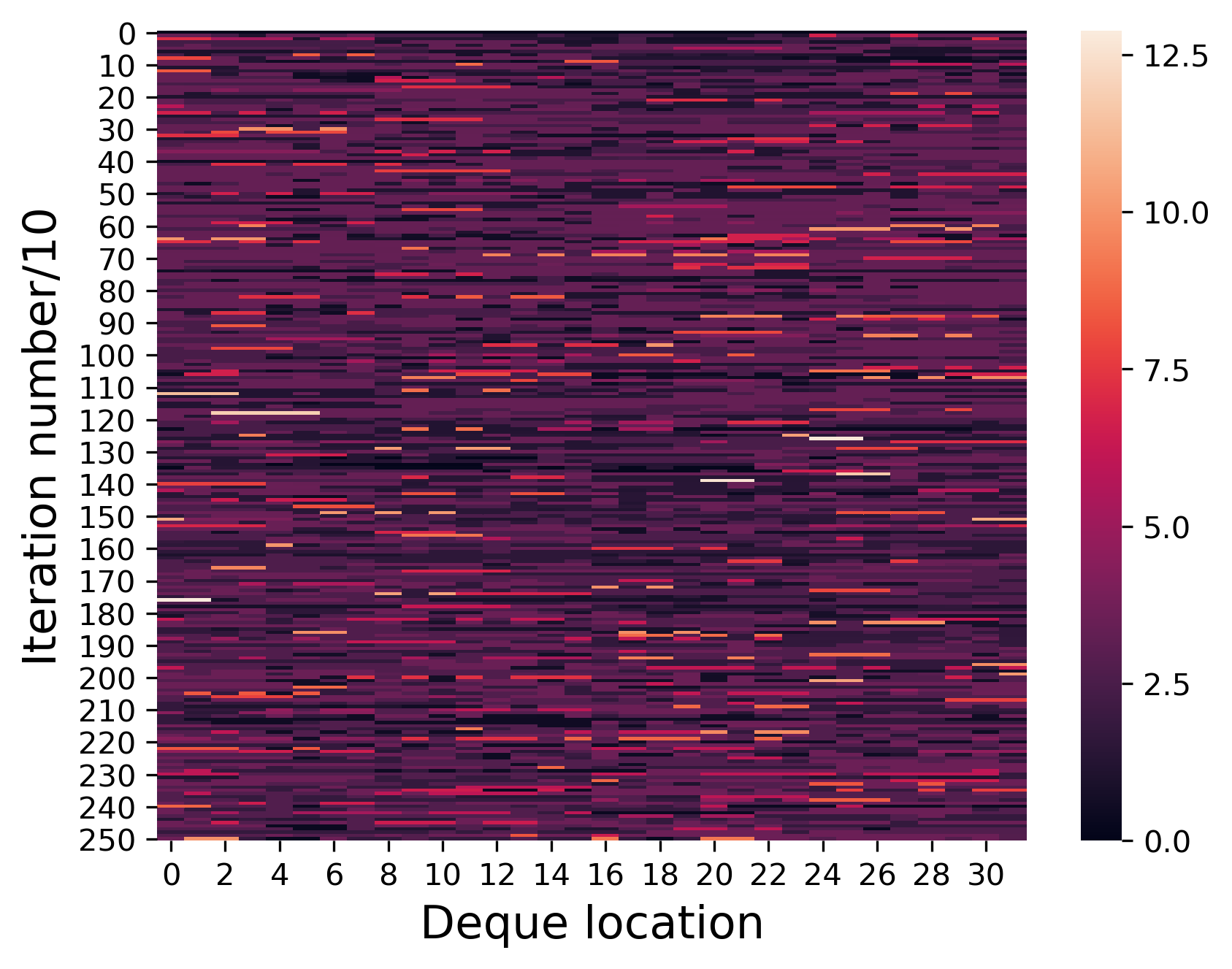}}}
\caption{Heatmap for trajectory priority with respect to memory size.The vertical axis indicates the number of training iterations; the horizontal axis indicates the position of the trajectory in the queue where the smaller the number is, the longer the trajectory generation time. The brighter the color in the graph, the higher the priority indicated.}
\label{fig:heatmap memory}
\end{figure*}

\begin{figure*}[htbp]
\centering
\subfigure[Max trajectory priority ]{\label{Max_length}{\includegraphics[width=4.5cm,height=4cm]{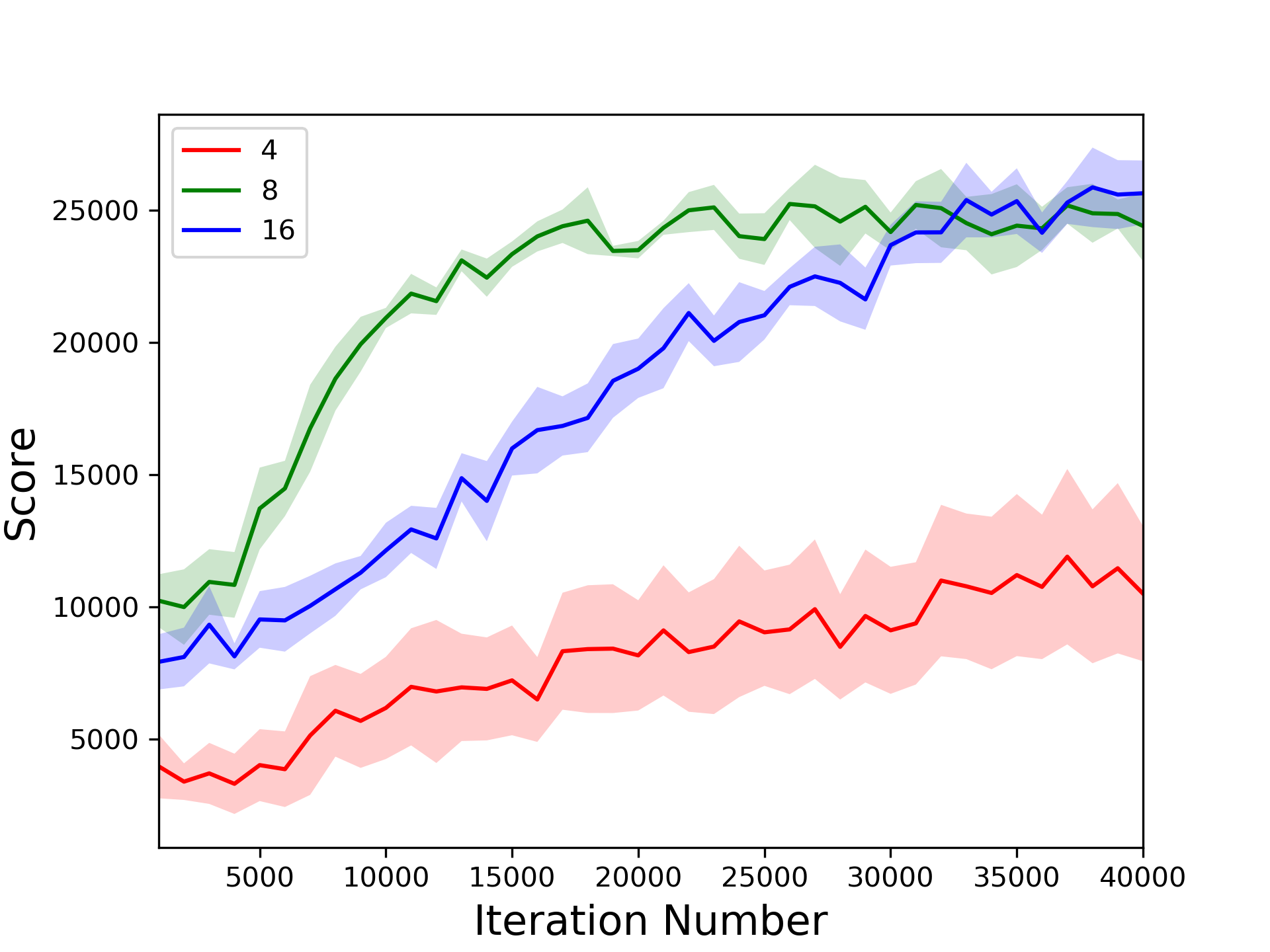}}}\hspace{1mm}
\subfigure[Mean trajectory priority ]{\label{Mean_length}{\includegraphics[width=4.5cm,height=4cm]{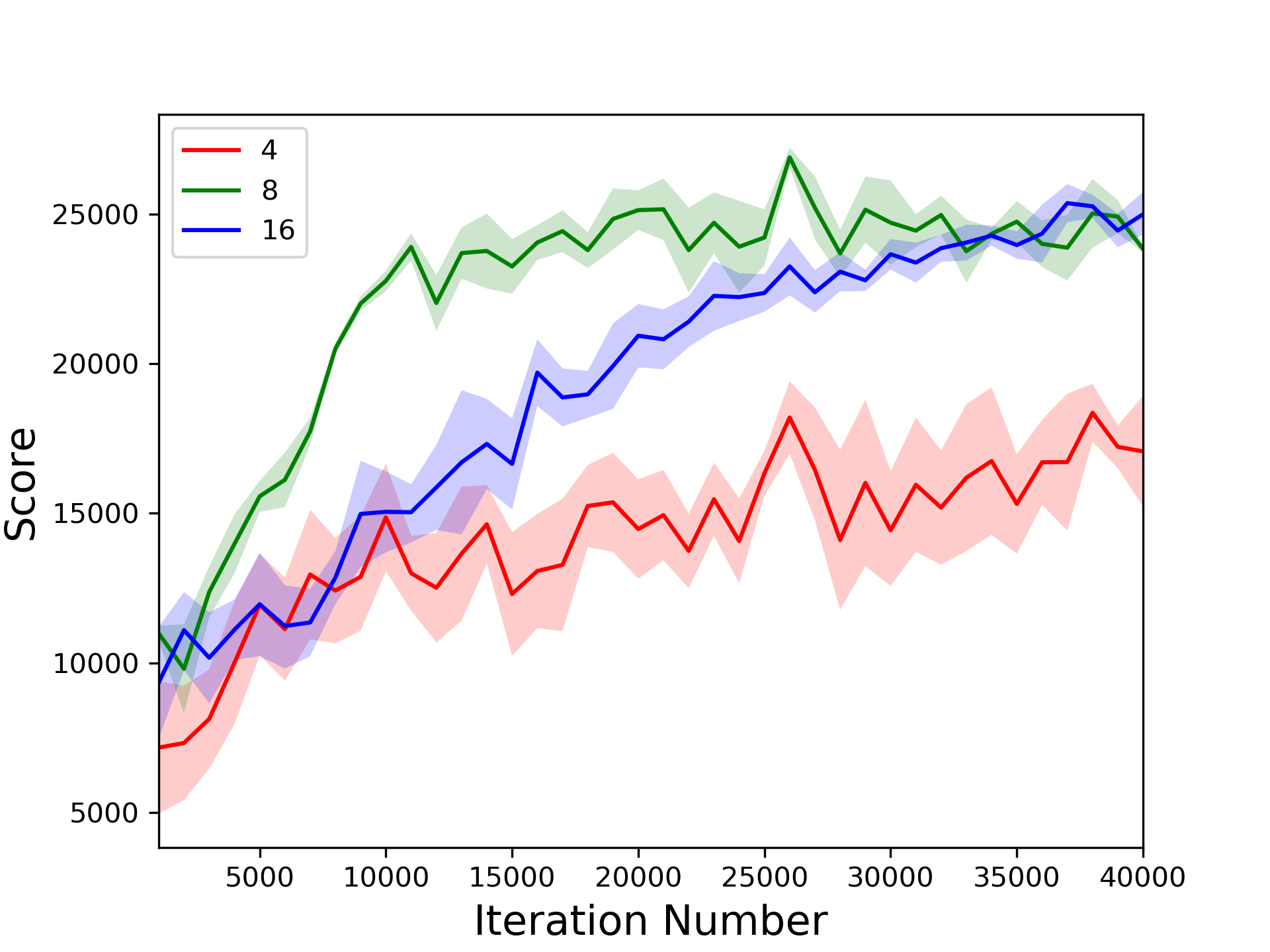}}}
\subfigure[Reward trajectory priority]{\label{Rew_length}{\includegraphics[width=4.5cm,height=4cm]{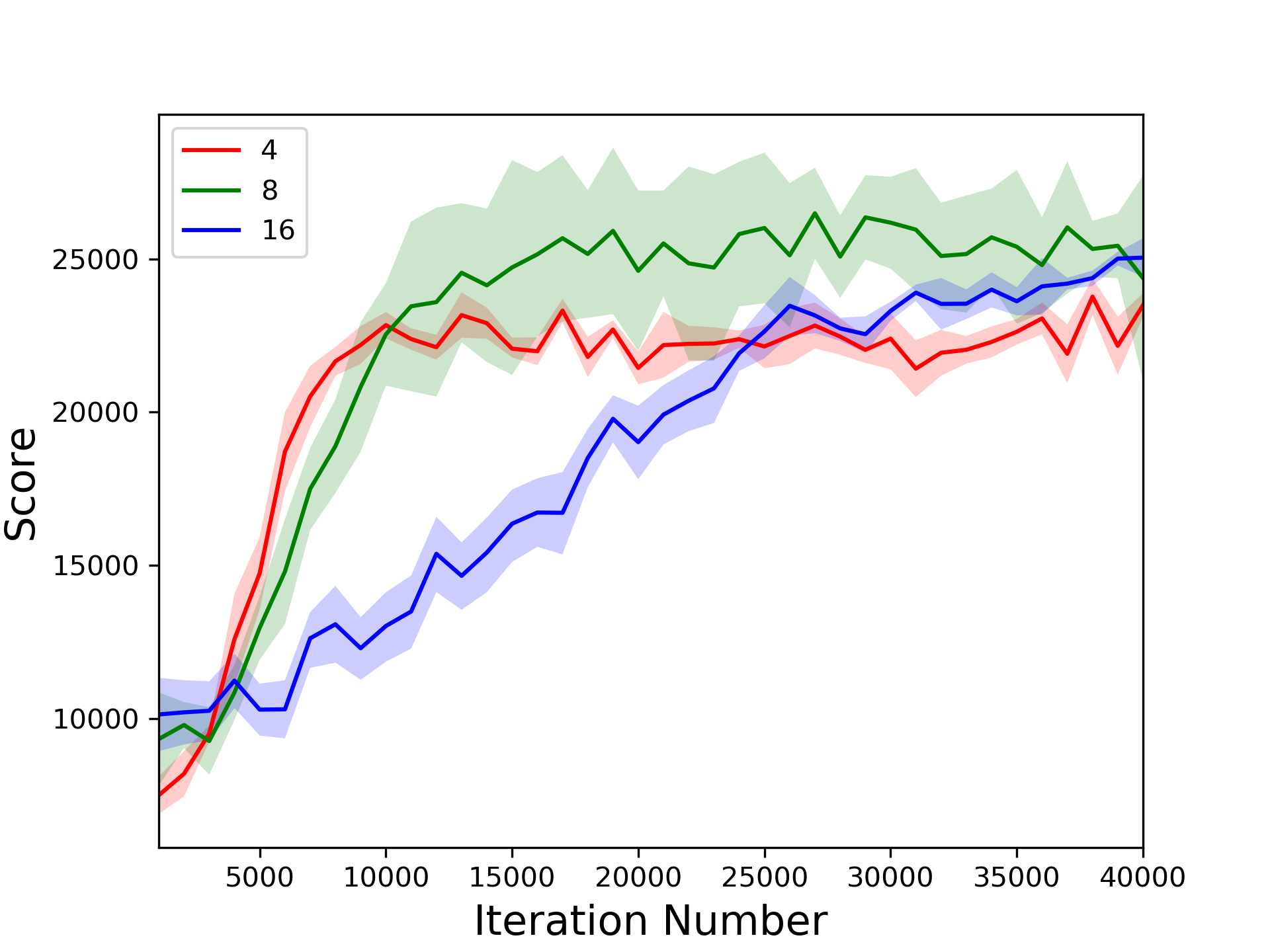}}}
\caption{Trajectory rollout length comparison. The red line indicates a rollout length of 4, the green line indicates a rollout length of 8, and the blue line indicates a rollout length of 16. (a) Comparison graph for the max trajectory priority, (b) comparison graph for the mean trajectory priority, and (c) comparison graph for the reward trajectory priority.}
\label{fig:length compare}
\centering
\subfigure[rollout size of max priority= 16 ]{\label{max_step16}{\includegraphics[width=4.5cm,height=4cm]{max_16_1024_8.png}}}\hspace{1mm}
\subfigure[rollout size of max priority= 8 ]{\label{max_step8}{\includegraphics[width=4.5cm,height=4cm]{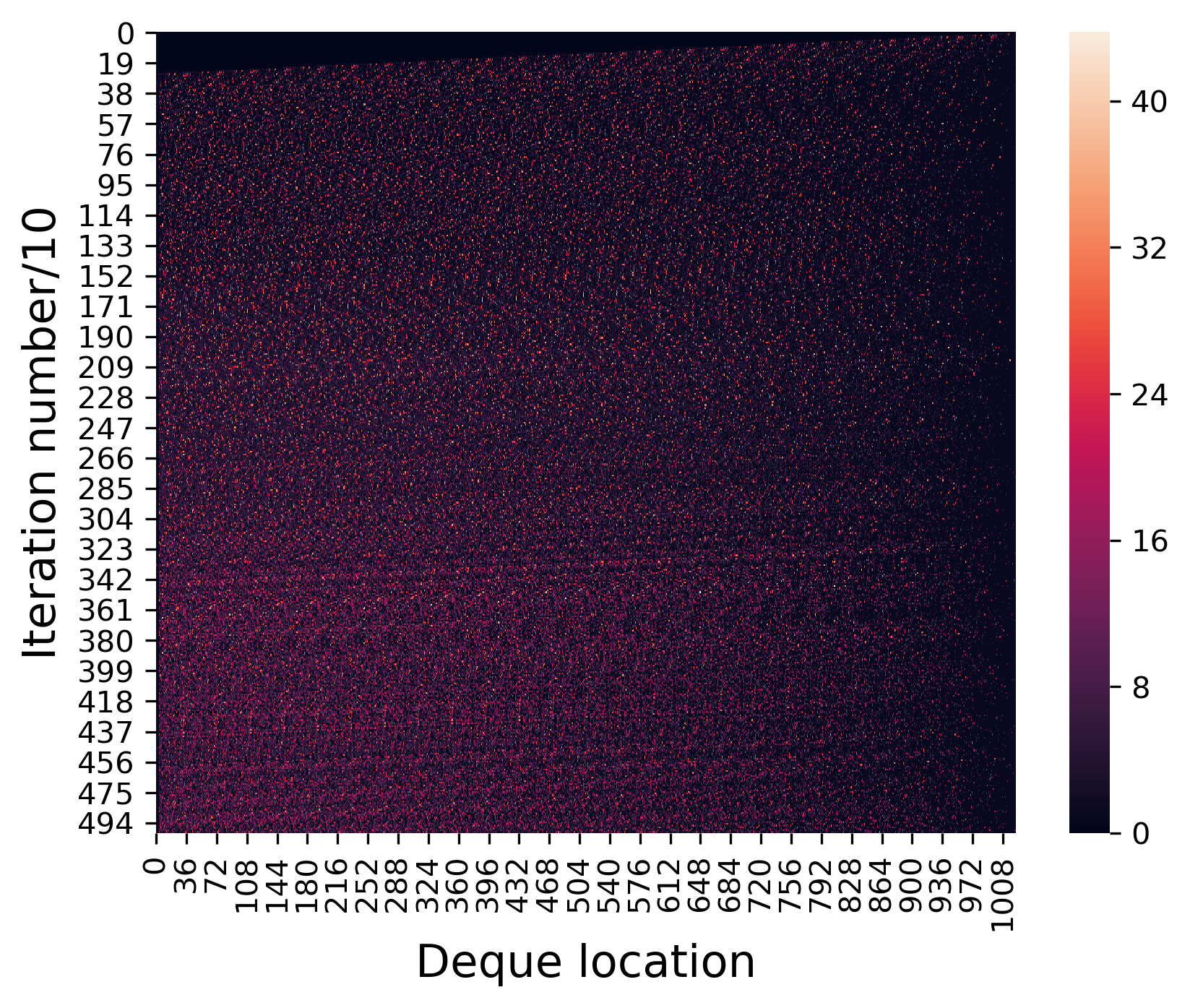}}}
\subfigure[rollout size of max priority= 4]{\label{max_step4}{\includegraphics[width=4.5cm,height=4cm]{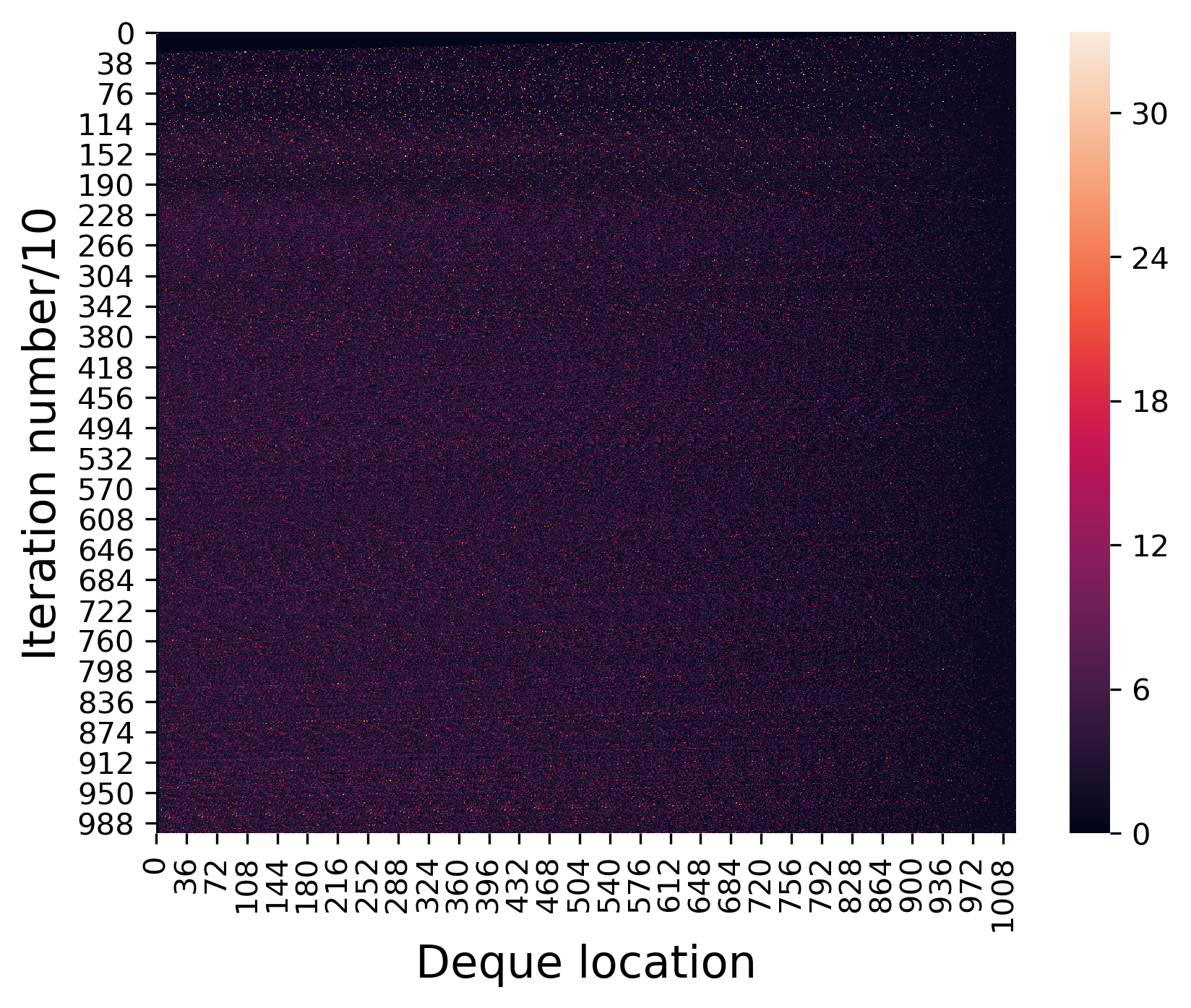}}}
\subfigure[rollout size of mean priority = 16 ]{\label{mean_step16}{\includegraphics[width=4.5cm,height=4cm]{mean_16_1024_8.png}}}\hspace{1mm}
\subfigure[rollout size of mean priority= 8 ]{\label{mean_step8}{\includegraphics[width=4.5cm,height=4cm]{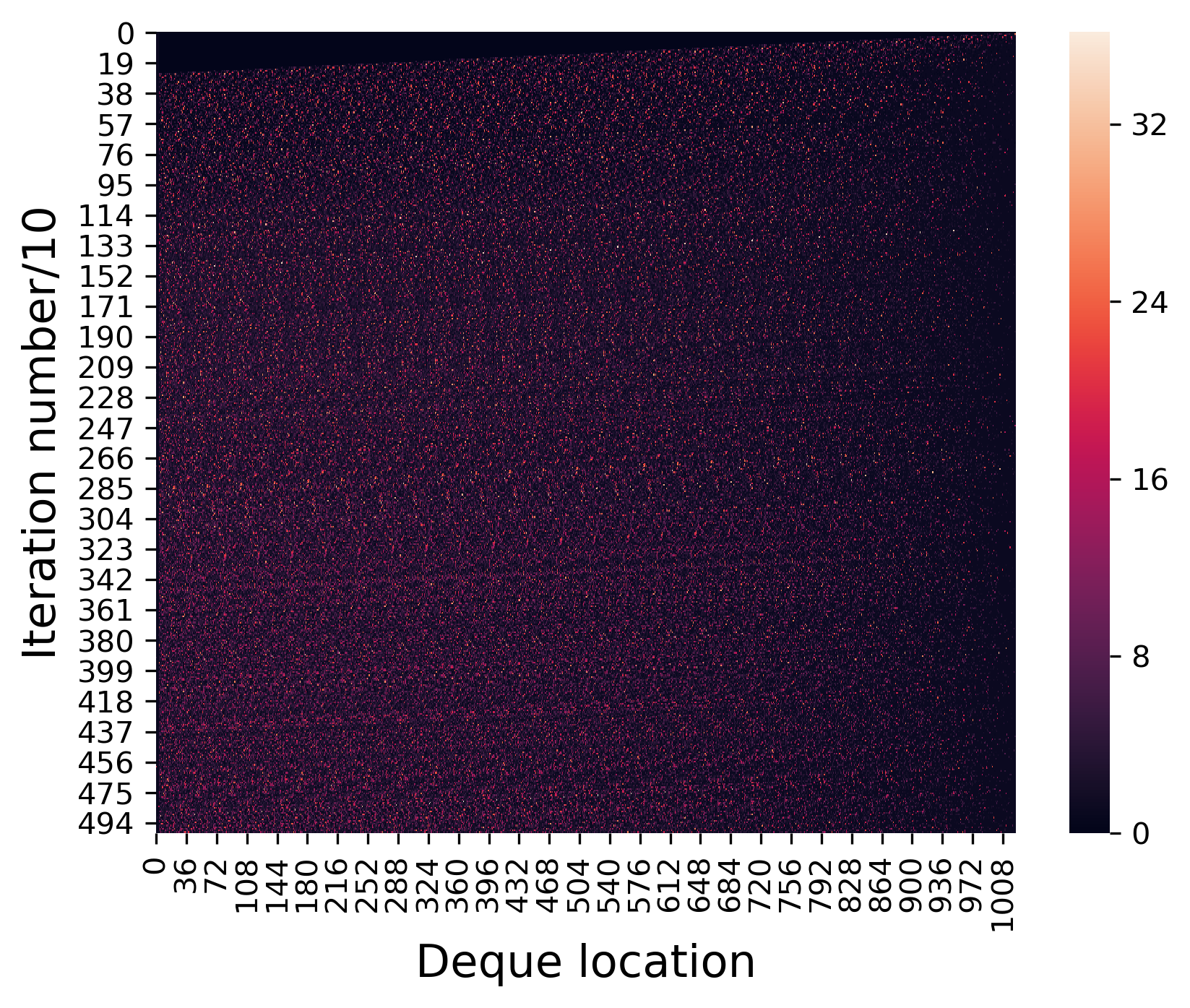}}}
\subfigure[rollout size of mean priority= 4]{\label{mean_step4}{\includegraphics[width=4.5cm,height=4cm]{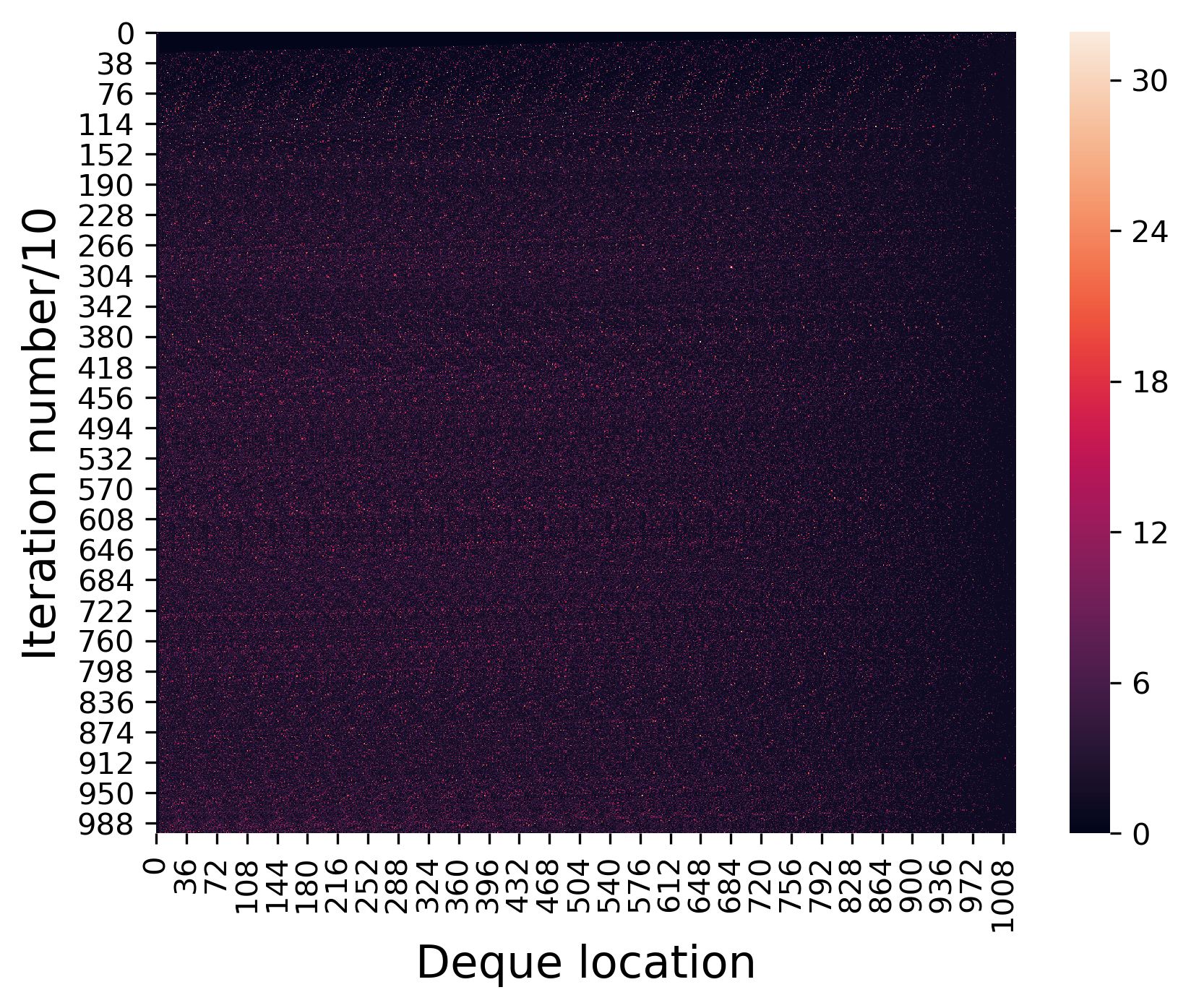}}}
\subfigure[rollout size of reward priority= 16 ]{\label{rew_step16}{\includegraphics[width=4.5cm,height=4cm]{rew_16_1024_8.png}}}\hspace{1mm}
\subfigure[rollout size of reward priority= 8 ]{\label{rew_step8}{\includegraphics[width=4.5cm,height=4cm]{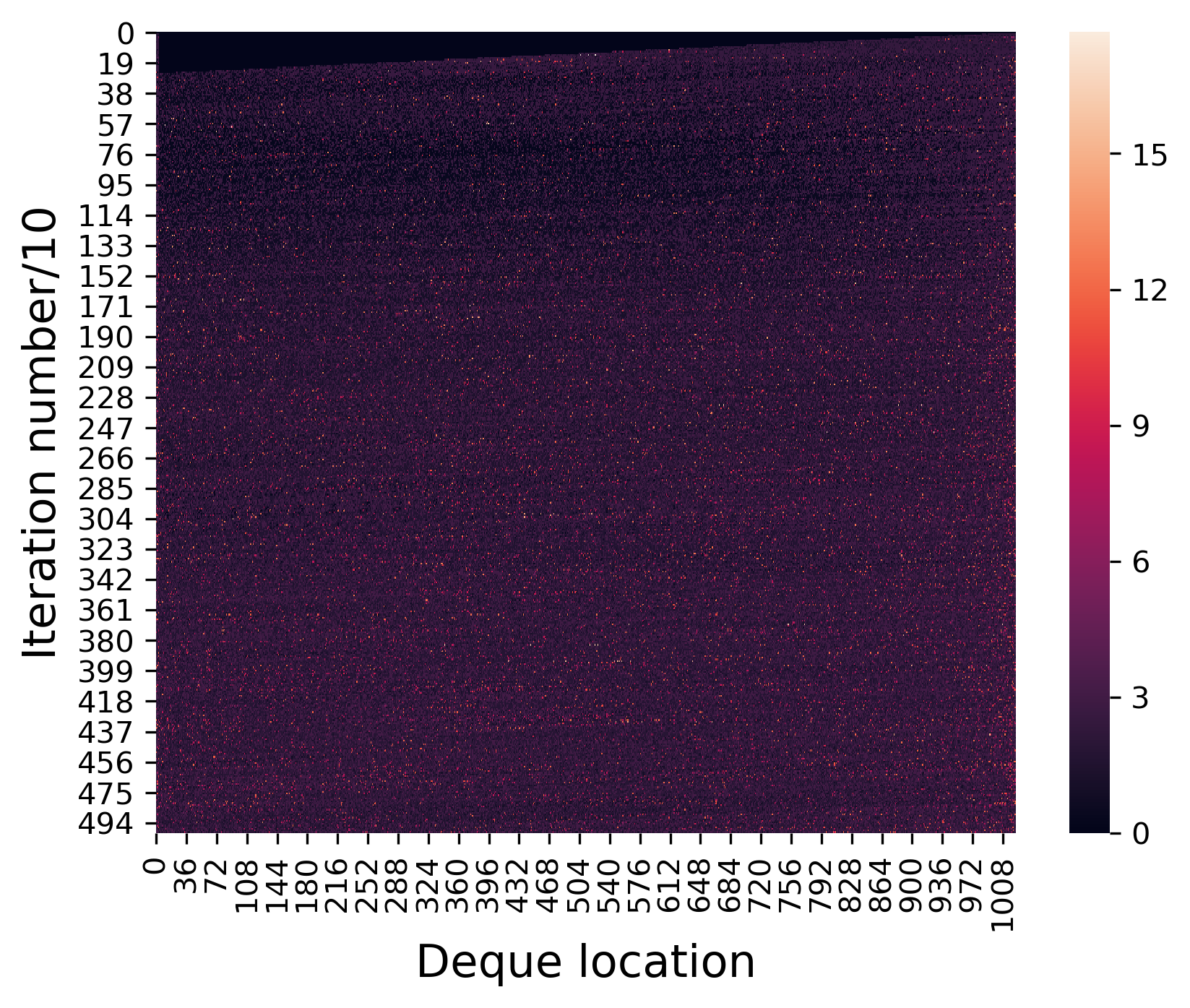}}}
\subfigure[rollout size of reward priority = 4]{\label{rew_step4}{\includegraphics[width=4.5cm,height=4cm]{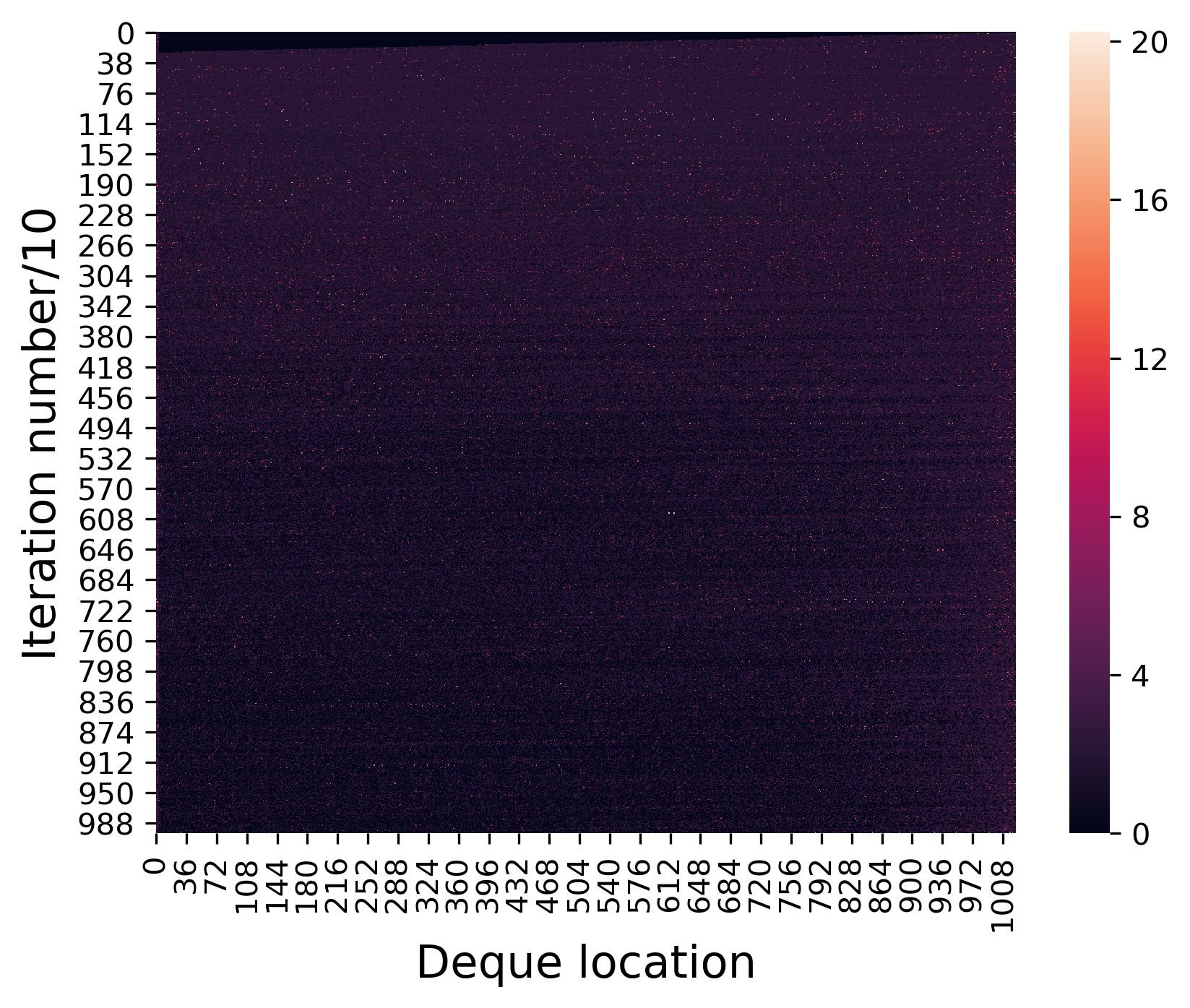}}}
\caption{Heatmap for trajectory priority with respect to rollout size. The vertical axis indicates the number of training iterations; the horizontal axis indicates the position of the trajectory in the queue. The smaller the number is, the longer the trajectory generation time. The brighter the color in the graph, the higher the priority indicated.}
\label{fig:heatmap rollout}
\end{figure*}

\subsubsection{Priority memory size comparison}
We examined the performance of the algorithm in the Atlantis-v0 environment with different priority memory sizes (32,256,1024) and a rollout length of 16 with max, mean, and reward trajectory priority. As shown in Fig. \ref{fig:memory compare}, the best performance and fastest learning were achieved when the memory size was 256, followed by a memory size of  1024 and a memory size of 32. Furthermore, at 40,000 steps, the final values for memory sizes of 256 and 1024 are were similar, both reaching near 25,000, while memory size 32 only learns to 15,000. 

To analyze the causes of these phenomena, we output the heatmaps of priority for different memory sizes, as shown in Fig. \ref{fig:heatmap memory}.

The subplots \ref{max_memory1024}-\ref{max_memory32} and \ref{mean_memory1024}-\ref{mean_memory32} show heatmaps of the trajectory priority for max and mean trajectory priority with memory sizes of 1024, 256 and 32. The heatmap trend for mean trajectory priority is similar to that of max trajectory priority. For a memory size of 1024, the high-priority trajectories were mostly distributed among the earlier trajectories due to the long-established network updates, while for a memory size of 32, all the recent trajectories were sampled back and forth. Each trajectory would be sampled multiple times in each update step, making the policy learning fall into local adjustment and slow learning. For a memory size of 256, the intensity of sample diversity was greater, avoiding the interference of multiple learning of recent trajectories, thereby preventing the large bias of the early trajectory estimation from interfering with trajectory prioritization. Thus, the memory size of 256 works best.

Subplots \ref{rew_memory1024}, \ref{rew_memory256} and \ref{rew_memory32} show the heatmaps of the trajectory priority for reward trajectory priority with memory sizes of 1024, 256 and 32. Subplot \ref{rew_memory1024} shows the heatmap at a memory size of 1024. In the early stages of training, the bright spots are concentrated on the newly acquired trajectories but do not differ much in brightness from other locations, and the sampling tends to uniform sampling. As the number of training sessions increases, the bright spots become more numerous and evenly distributed over all locations. Subplot \ref{rew_memory256} shows the heatmap at a memory size of 256, with significantly more bright spots than that in \ref{rew_memory1024}, with more bright spots and a wide distribution at all stages of training. Subplot \ref{rew_memory32} shows the heatmap at a memory size of 32, with significantly more bright spots present at all periods, significantly outperforming and dominating the other trajectories. The heatmap indicates that when the memory size is 1024, the priority on the games in the trajectory is similar and the differentiation is not obvious: the differentiation gradually increases as the training progresses. When the memory size is 256, there are more obvious bright spots with good differentiation at all stages of training; and when the memory size is 32, there are individual bright spots with high brightness at all periods, which are sampled in multiple times for training, thereby reducing the diversity of the trajectories. Thus, the best performance is achieved when the memory size is set to 256.

\subsubsection{Trajectory rollout length comparison}

Finally, we set the memory size to 1024 and tested the effect of different trajectory rollout lengths (4,8,16) on the PTR-PPO algorithm in the Atlantis-v0 environment. As shown in Fig.\ref{fig:length compare}, the algorithm performs best when the rollout length is set to 8, second best when it is set to 16. In reward trajectory priority, the learning speed is similar when the rollout length is set to 4 and 8, and the slowest when the length is 16, but the final value is best for length 8, second best for 16 and worst for length 4. 

Again, to analyze the causes of this phenomenon, the historical values of the three priorities were recorded and presented in the form of a heatmap, as shown in Fig.\ref{fig:heatmap rollout} .

Comparing \ref{max_step16}, \ref{max_step8} and \ref{max_step4}, we find that \ref{max_step8} has an obvious bright spot demarcation when the vertical coordinate is 247, with a clear distribution of trajectory priority differences in the upper part and weak trajectory priority differences in the lower part. Subplot \ref{max_step4} has an obvious demarcation when the vertical coordinate is 190, while subplot \ref{max_step16} has no obvious demarcation, with the trajectory priority at all stages being significantly different. These demarcation points correspond to different learning stages in Fig. \ref{Max_length}, with length 16 always in the climbing stage, while length 8, at 22000 steps, enters a smooth stage; and length 4, at 8000 steps, has a clear climb and then falls into a slow learning improvement stage.

Comparing subplots \ref{mean_step16}, \ref{mean_step8} and \ref{mean_step4}, we find that subplot \ref{mean_step8} has a clear bright spot demarcation at a vertical coordinate of 309, with a clear distribution of trajectory priority differences in the upper part and weaker trajectory priority differences in the lower part. Subplot \ref{mean_step4} has a clear demarcation at a vertical coordinate of 228, and subplot \ref{mean_step16} has a dark section between 230 and 250. Again, these demarcations correspond to different learning stages in Fig. \ref{Mean_length}, with length 16 entering a plateau at 35,000 steps, length 8 reaching a plateau at 20,000 steps, and length 4 having a clear climb until 10,000 steps and then falling into a slow learning improvement phase.

Subplots \ref{max_step16}-\ref{mean_step4} are all GAE-based trajectory priorities with the same learning characteristics, and the degree of weight differentiation of the trajectories in memory determines whether the algorithm is in a learning improvement or a smooth learning phase.

Compared to the six subgraphs above, subgraphs \ref{rew_step16}-\ref{rew_step4} have different characteristics. As the rollout length is adjusted from 16 to 4, the number of bright spots in the three subplots gradually decreases. Subplot \ref{rew_step16} has a large number of bright spots at each stage, most of which are significant trajectories. By contrast, subplot \ref{rew_step4}, at different stages, has a small number of unevenly distributed bright spots at different locations, while subplot \ref{rew_step8} falls in between, with the number of bright spots at each stage at an appropriate value. Thus, in the reward trajectory priority, a rollout length that creates a good degree of differentiation in memory works best.

\section{Conclusion}

To improve the sample efficiency of on-policy deep reinforcement learning methods, this paper proposes a novel PPO algorithm with prioritized trajectory replay (PTR-PPO), which improves the performance of the PPO algorithm by reusing trajectories with high priority. We first define three trajectory priorities, max and mean trajectory priorities based on GAE, and reward trajectory priorities based on normalized undiscounted cumulative reward. Then, to overcome the high variance caused by the large importance weights, we propose an importance weight estimation method with truncation to define the policy improvement loss function for PPO under off-policy conditions. We evaluate PTR-PPO and other baseline algorithms in a set of Atari tasks. The experimental results show that the PTR-PPO algorithm outperforms the other baseline algorithms.

The three factors have differences in the trajectory weights in the priority memory but share the same network architecture and can share parameters. In future work, we will design an integrated architecture where the three factors are designed as a hyperparameter to call different priority memories at different stages for policy improvement.

\bibliographystyle{unsrtnat}
\bibliography{references}  

\end{document}